%% file: acl_latex.tex
\pdfoutput=1

\documentclass[11pt]{article}

\usepackage[preprint]{acl}

\usepackage{times}
\usepackage{latexsym}
\usepackage{algorithm}
\usepackage{algorithmic}
\usepackage{multirow}
\usepackage{enumitem}
\usepackage{booktabs}
\usepackage{amsmath}
\usepackage{amssymb}
\usepackage{subcaption}
\usepackage[most]{tcolorbox}
\usepackage{xcolor}
\usepackage{times}
\usepackage{helvet}
\usepackage{courier}
\usepackage{amsfonts,bm}
\usepackage{stfloats}

\usepackage[T1]{fontenc}

\usepackage[utf8]{inputenc}

\usepackage{microtype}

\usepackage{inconsolata}

\usepackage{graphicx}

\title{Thinking with Nothinking Calibration: A New In-Context Learning Paradigm in Reasoning Large Language Models}

\author{%
    Haotian Wu$^\dagger$, 
    Bo Xu$^\dagger$, 
    Yao Shu$^\dagger$, 
    Menglin Yang$^\dagger$, 
    Chengwei Qin$^\dagger$ \\
    $^\dagger$The Hong Kong University of Science and Technology (Guangzhou) \\
    \{markhaotianw, bx, yaoshu, menglinyang, chengweiqin\}@hkust-gz.edu.cn
}

\begin{document}
\maketitle
\input{./sections/abstract}
\input{./sections/introduction}
\input{./sections/related_work}
\input{./sections/method}
\input{./sections/experiments}
\input{./sections/analysis}
\input{./sections/conclusion}
\input{./sections/limitations}
\bibliography{custom}
\input{./sections/appendix}
\end{document}

%% file: sections/abstract.tex
\begin{abstract}
Reasoning large language models (RLLMs) have recently demonstrated remarkable capabilities through structured and multi-step reasoning. While prior research has primarily focused on improving their training and inference strategies, their potential for in-context learning (ICL) remains largely underexplored. To fill this gap, we propose \textit{Thinking with Nothinking Calibration (JointThinking)}, a new ICL paradigm that prompts the model to generate two answers in parallel: one in \textit{Thinking} mode and the other in \textit{Nothinking} mode. A second round of \textit{Thinking} is triggered only when the two initial responses are inconsistent, using a single prompt with two different answers.
Extensive experiments across multiple reasoning benchmarks demonstrate that \textit{JointThinking} significantly outperforms few-shot chain-of-thought (CoT), thinking twice and majority voting.
Moreover, it achieves comparable in-distribution performance to training-based SOTA reasoning method, while substantially outperforming on out-of-distribution tasks. 
We further conduct a systematic analysis of the calibration mechanism, showing the importance of structural thinking diversity and the benefits of consistency check.
Additionally, we observe that the performance gap between actual and ideal reasoning narrows as model size increases in the second thinking, indicating the strong scalability of our approach. Finally, we discuss current limitations and outline promising directions for future ICL research in RLLMs. Our codes are available at \url{https://github.com/HAOTIAN89/JointThinking}.
\end{abstract}

%% file: sections/introduction.tex
\section{Introduction}


Reasoning large language models (RLLMs), such as OpenAI-o3 \cite{OpenAI2024IntroducingO3}, DeepSeek-R1 \cite{guo2025deepseek}, and Qwen3 \cite{yang2025qwen3}, mark a fundamental shift in model output paradigms by explicitly structuring the inference process. These models significantly improve performance on complex reasoning tasks by increasing inference-time compute \cite{snell2024scaling}. The inference process of RLLMs typically consists of two stages: a "Thinking" phase where the model generates reasoning chains through iterative exploration, reflection, and verification; and a "Final Solution" phase presenting the refined reasoning and answer. Such structured reasoning behavior is often acquired through reinforcement learning (RL), as noted by \cite{guo2025deepseek}. Despite their strong reasoning capabilities, recent studies have identified a phenomenon known as ``Overthinking'' \cite{cuadron2025danger,chen2024not}, where RLLMs continue reasoning beyond correct solutions due to uncertainty, adding unnecessary complexity and errors, particularly on simpler problems where traditional large language models (LLMs) may perform better.

\begin{figure}[t]
    \includegraphics[width=0.95\linewidth]{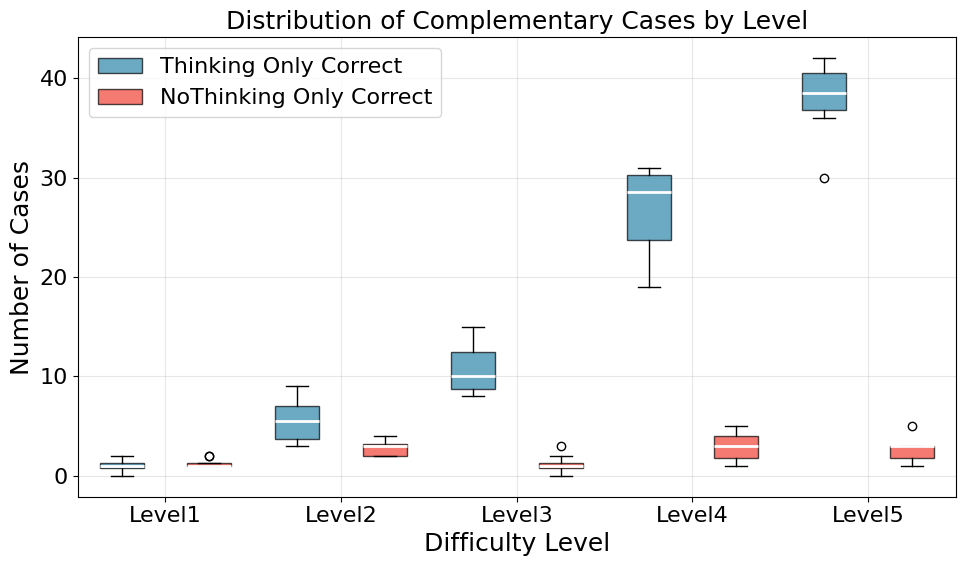}
    \caption{Comparison of R1-7B using \textit{Thinking} and \textit{Nothinking} on MATH500. Each mode reliably compensates for the other’s failures across all difficulty levels.}
    \label{complementary}
\end{figure}


\begin{figure*}[htp!]
  \centering
  \includegraphics[width=.95\linewidth]{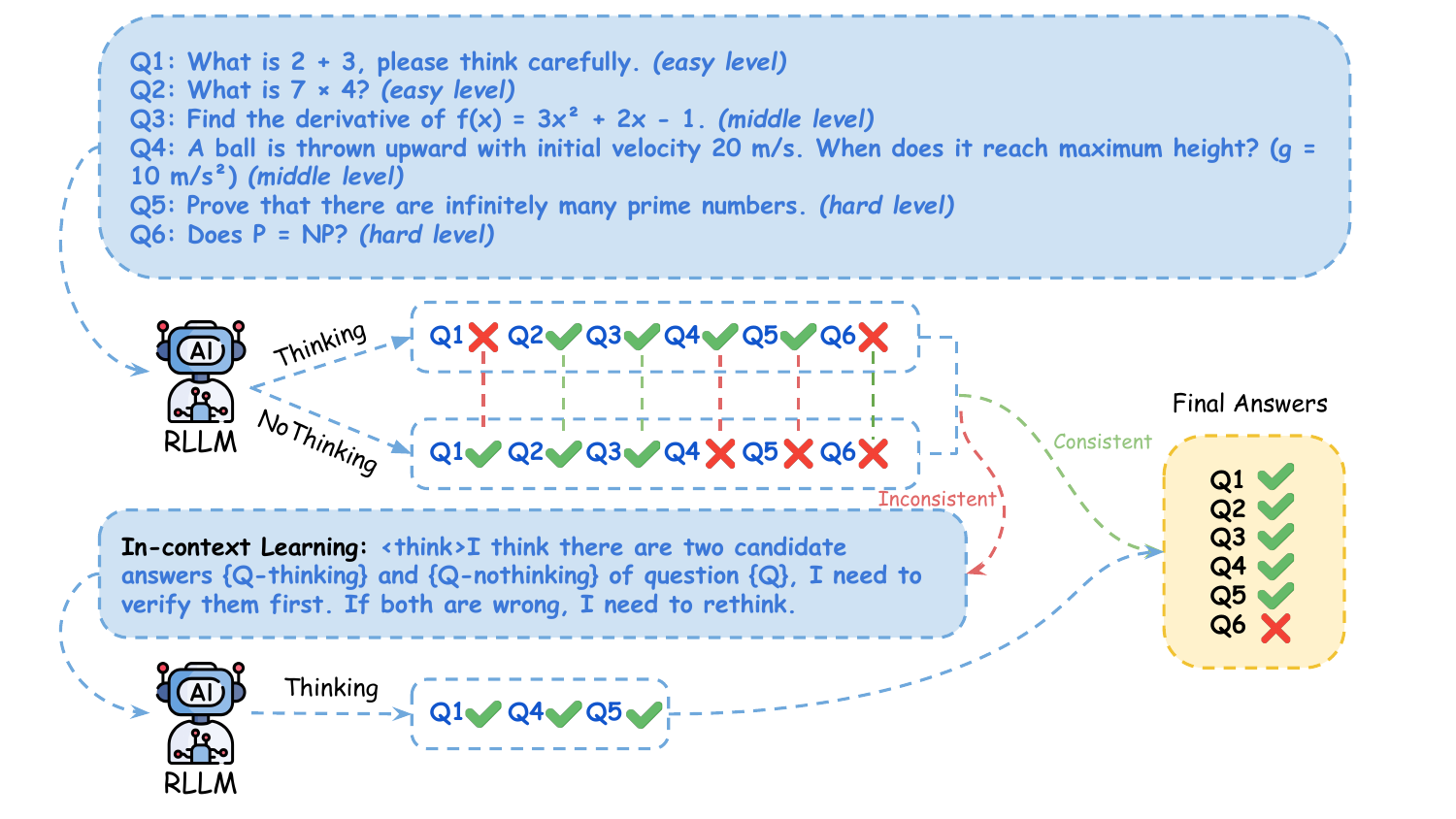}
  \caption{\textit{Thinking with Nothinking Calibration (JointThinking)}. Given one question, the reasoning large language model generates two outputs in parallel with \textit{Thinking} and \textit{Nothinking} modes. If answers are inconsistent, the model will perform a second round of reasoning in the \textit{Thinking} mode, using the two inconsistent answers as context. 
  }
  \label{fig:framework}
\end{figure*}

In the broader LLM landscape, in-context learning (ICL) remains a highly active research area due to its flexibility and effectiveness. While prior work on RLLMs has primarily focused on improving training and inference strategies, relatively little attention has been paid to their ICL capabilities. A recent study by \cite{ge2025innate} investigates the effects of Zero-CoT and Few-shot CoT on RLLMs, suggesting that few-shot CoT improves reasoning performance. However, our empirical results reveal that the improvement is limited. We hypothesize that RL enhances the model's intrinsic ability to discover optimal reasoning paths, thereby reducing reliance on explicit prompt-based guidance. Indeed, recent work \cite{li2025thinking} has also shown that explicit CoT reasoning can sometimes degrade instruction-following accuracy by omitting the constraints of instruction in the prompt.


Given these observations, we aim to design an ICL paradigm tailored to the strengths of RLLMs, i.e., their flexible and adaptive reasoning capabilities, while minimizing reliance on prompt design. The \textit{Thinking} mode performs well on complex problems but is prone to overthinking and making mistakes on simpler ones. In contrast, the \textit{Nothinking} mode \cite{ma2025reasoning}, while less effective on challenging questions, achieves higher accuracy on easier ones \cite{zhang2025adaptthink,lou2025adacot}. Through an initial analysis of the MATH500 dataset in Figure \ref{complementary}, we identify a complementary pattern that each mode reliably compensates for the other's failures across all difficulty levels.

Building on these insights, we propose a novel ICL framework for RLLMs called \emph{Thinking with Nothinking Calibration} (or \emph{JointThinking}). Given a question, the model generates two responses in parallel: one using the \emph{Thinking} mode and the other using the \emph{Nothinking} mode. If both responses are consistent, the answer from the \textit{Thinking} mode is adopted. Otherwise, both answers are fed into the prompt for a second round of \emph{Thinking} to resolve the disagreement (see Figure \ref{fig:framework}). Extensive experiments show that our method significantly outperforms single-mode reasoning, few-shot CoT, thinking twice and majority voting baselines across various models and mathematical benchmarks. Moreover, it achieves performance comparable to training-based state-of-the-art (SOTA) reasoning enhancement method under in-distribution tasks, while demonstrating remarkable superiority in out-of-distribution scenarios.

We further validate the necessity of incorporating different reasoning modes and the consistency check mechanism. Mixed-mode calibration consistently yields lower error rates than using a single mode, while consistency check to selectively trigger second thinking not only reduces computational overhead but can also achieve better performance than always performing second thinking. Additionally, we find a clear scaling trend in the second-thinking stage: as model size increases, the gap between the generated answers and the ideal solutions progressively narrows, with our method in some cases even surpassing the ideal calibration performance. Finally, we explore current limitations of ICL in RLLMs and highlight promising future research directions. In summary, our main contributions are:


\begin{itemize}[leftmargin=*,labelsep=2mm]
    \item We introduce \emph{Thinking with Nothinking Calibration} (or \emph{JointThinking}), a novel ICL paradigm for RLLMs that leverages the complementary strengths of \emph{Thinking} and \emph{Nothinking} modes.
    \item With extensive experiments, we demonstrate that \emph{JointThinking} significantly outperforms existing baseline approaches and exhibits strong generalization to out-of-distribution scenarios.
    \item We systematically analyze the benefits of mixed-mode calibration and the consistency check mechanism. We further investigate how model scaling influences second-thinking performance.
\end{itemize}

%% file: sections/related_work.tex
\section{Related Work}

\paragraph{In-Context Learning}
In-context learning (ICL) was first proposed by \cite{brown2020languagemodelsfewshotlearners}, which refers to the capability of large language models (LLMs) to perform tasks solely by being provided with examples or instructions within the input prompt, without any updates to their internal parameters. The ICL capabilities of LLMs have been shown to improve further through self-supervised or supervised training \cite{wei2023symbol}. Since then, an increasing number of studies have focused on this simple yet effective paradigm. Some works have studied how to design good demonstrations \cite{tang2024demonstrationnotebookfindingsuited,peng-etal-2024-revisiting,wang2025demonstrationselectionincontextlearning} and how to build effective ICL-based applications \cite{meade2023using,buoso2024select2plantrainingfreeiclbasedplanning}. Other works have focused on theoretical perspectives to analyze why ICL works for LLMs. \cite{wies2023learnability} establishes a PAC-based theoretical framework for ICL, showing that in-context learning primarily involves task identification rather than task learning. \cite{zhang2025promptdesignmattersworks} shows that each prompt defines a unique trajectory through the answer space, and the choice of trajectory is crucial for task performance and future navigation within the space during CoT reasoning. Models of different scales exhibit distinct ICL behaviors, with smaller models being more robust to noise while larger models tend to over-attend to irrelevant information. \cite{shi2024largerlanguagemodelsincontext}.

\paragraph{Reasoning Large Language Models}
The emergence of test-time scaling RL methods, such as GRPO \cite{shao2024deepseekmath}, significantly enhances the reasoning capabilities of large language models for complex problems, especially mathematical problems. These trained models are referred to as reasoning large language models (RLLMs) \cite{guo2025deepseek,yang2025qwen3}, which first generate an extensive chain of thought (CoT), often involving techniques such as backtracking and verification, before arriving at a final explanation and answer. For the training stage, many studies explore how to effectively train models with suitable-length thinking processes, including reinforcement learning with length reward design \cite{zhang2025adaptthink,aggarwal2025l1,lou2025adacot} and supervised fine-tuning with variable-length CoT data \cite{yu2024distilling,ma2025cot,yu2025z1}. Works exploring the inference stage mainly focus on sampling multiple outputs and aggregating them in parallel, including Best-of-N related methods \cite{sun2024fast,wang2025sampling,ma2025reasoning} and search-guided related methods \cite{li2025fastmcts,liao2025reward}. Additionally, \cite{ge2025innate} conducts a comprehensive analysis examining the impact of few-shot CoT on RLLMs from the ICL perspective, demonstrating its positive influence on reasoning performance.


%% file: sections/method.tex
\begin{table*}[h]
    \centering
    \scalebox{0.95}{
    \renewcommand{\arraystretch}{0.9}
    {\small
    \begin{tabular}{lcccccc}
    \toprule
    \multicolumn{1}{c}{\multirow{2}{*}{\textbf{Model}}} & \multicolumn{1}{c}{\multirow{2}{*}{\textbf{Method}}} & \multicolumn{4}{c}{\textbf{Datasets (Pass@1)}} &
    \multicolumn{1}{c}{\multirow{2}{*}{\textbf{Average}}} \\
    \cmidrule{3-6}
    \multicolumn{1}{c}{} & \multicolumn{1}{c}{} & GSM8K(1319) & MATH500(500) & AIME24(30) & AMC23(40) & \multicolumn{1}{c}{} \\
    \midrule
    \multicolumn{1}{c}{\multirow{5}{*}{R1-1.5B}} & Thinking & 81.80 & 78.20 & 23.33 & 66.25 & 62.40 \\
    \multicolumn{1}{c}{} & Nothinking & 77.94 & 79.00 & 19.17 & 68.75 & 61.22 \\
    \multicolumn{1}{c}{} & Few-shot & 81.73 & 71.20 & 22.08 & 54.69 & 57.43 \\
    \multicolumn{1}{c}{} & Majority Voting & \underline{83.93} & \underline{81.60} & 26.67 & \underline{70.62} & \underline{65.71} \\
    \multicolumn{1}{c}{} & Thinking Twice & 82.87 & 78.00 & \underline{30.83} & 70.31 & 65.50 \\
    \multicolumn{1}{c}{} & JointThinking (\textbf{Ours}) & \textbf{84.23} & \textbf{82.80} & \textbf{35.00} & \textbf{72.41} & \textbf{68.61} \\
    \midrule
    \multicolumn{1}{c}{\multirow{5}{*}{R1-7B}} & Thinking & 87.79 & 87.60 & 47.50 & 86.88 & 77.44 \\
    \multicolumn{1}{c}{} & Nothinking & 85.90 & 74.60 & 22.92 & 59.38 & 60.70 \\
    \multicolumn{1}{c}{} & Few-shot & 84.91 & 87.00 & 35.83 & 73.13 & 70.22 \\
    \multicolumn{1}{c}{} & Majority Voting & 89.99 & \textbf{88.00} & 50.00 & \underline{87.50} & \underline{78.87} \\
    \multicolumn{1}{c}{} & Thinking Twice & \underline{90.07} & 87.00 & \underline{50.42} & 86.88 & 78.59 \\
    \multicolumn{1}{c}{} & JointThinking (\textbf{Ours}) & \textbf{91.05} & \underline{87.80} & \textbf{51.25} & \textbf{88.44} & \textbf{79.64} \\
    \midrule
    \multicolumn{1}{c}{\multirow{5}{*}{R1-14B}} & Thinking & 89.99 & 88.20 & 63.75 & 90.00 & 82.99 \\
    \multicolumn{1}{c}{} & Nothinking & 90.14 & 74.00 & 30.42 & 72.81 & 66.84 \\
    \multicolumn{1}{c}{} & Few-shot & 89.08 & 88.40 & 58.33 & 89.69 & 81.38 \\
    \multicolumn{1}{c}{} & Majority Voting & 91.66 & 89.00 & \underline{65.00} & 90.31 & \underline{83.99} \\
    \multicolumn{1}{c}{} & Thinking Twice & \underline{93.71} & \underline{89.20} & 60.00 & \underline{92.19} & 83.78 \\
    \multicolumn{1}{c}{} & JointThinking (\textbf{Ours}) & \textbf{93.93} & \textbf{89.80} & \textbf{65.42} & \textbf{93.75} & \textbf{85.73} \\
    \midrule
    \multicolumn{1}{c}{\multirow{5}{*}{R1-32B}} & Thinking & 92.80 & 86.20 & 65.42 & 92.50 & 84.23 \\
    \multicolumn{1}{c}{} & Nothinking & 92.72 & 78.20 & 39.58 & 68.97 & 69.87 \\
    \multicolumn{1}{c}{} & Few-shot & 90.98 & 88.20 & 53.75 & 90.62 & 80.89 \\
    \multicolumn{1}{c}{} & Majority Voting & 93.93 & 88.60 & 66.67 & 93.13 & 85.58 \\
    \multicolumn{1}{c}{} & Thinking Twice & \underline{95.07} & \underline{88.80} & \underline{69.58} & \underline{93.75} & \underline{86.80} \\
    \multicolumn{1}{c}{} & JointThinking (\textbf{Ours}) & \textbf{96.29} & \textbf{89.80} & \textbf{70.83} & \textbf{94.69} & \textbf{87.90} \\
    \midrule
    \multicolumn{1}{c}{\multirow{5}{*}{Qwen3-8B}} & Thinking & 94.69 & 87.80 & 59.17 & 86.56 & 82.06 \\
    \multicolumn{1}{c}{} & Nothinking & 92.19 & 80.80 & 29.58 & 73.36 & 68.98 \\
    \multicolumn{1}{c}{} & Few-shot & 94.47 & \underline{89.80} & 57.50 & 88.12 & 82.47 \\
    \multicolumn{1}{c}{} & Majority Voting & \underline{94.92} & 88.60 & \underline{65.00} & \underline{90.00} & \underline{84.63} \\
    \multicolumn{1}{c}{} & Thinking Twice & 94.62 & 89.60 & 64.58 & 88.75 & 84.39 \\
    \multicolumn{1}{c}{} & JointThinking (\textbf{Ours}) & \textbf{95.00} & \textbf{91.20} & \textbf{71.25} & \textbf{92.81} & \textbf{87.57} \\
    \midrule
    \multicolumn{1}{c}{\multirow{5}{*}{Qwen3-14B}} & Thinking & 94.77 & 89.80 & 62.92 & 92.50 & 85.00 \\
    \multicolumn{1}{c}{} & Nothinking & 92.49 & 84.00 & 27.50 & 74.76 & 69.69 \\
    \multicolumn{1}{c}{} & Few-shot & \underline{95.30} & 90.00 & 53.75 & 92.50 & 82.89 \\
    \multicolumn{1}{c}{} & Majority Voting & 95.22 & 90.20 & 65.00 & 92.50 & 85.73 \\
    \multicolumn{1}{c}{} & Thinking Twice & 94.77 & \underline{91.80} & \underline{70.00} & \underline{96.25} & \underline{88.21} \\
    \multicolumn{1}{c}{} & JointThinking (\textbf{Ours}) & \textbf{95.45} & \textbf{92.59} & \textbf{71.67} & \textbf{96.88} & \textbf{89.15} \\
    \bottomrule
    \end{tabular}}}
    \vspace{-0.3em}
    \caption{Accuracy (\%) of \emph{JointThinking} and no-training baselines across different models and benchmarks, and for AIME24 and AMC23 the results are averaged by 8 individual runs. The best and second results are \textbf{bolded} and \underline{underlined}, respectively. Prompt template of different methods can be found in Appendix \ref{appendix:prompt_details}.}
    \label{table:main results}
\end{table*}

\begin{table*}[t]
    \centering
    \scalebox{0.95}{\small
    \begin{tabular}{lcccccc}
    \toprule
    \multicolumn{1}{c}{\multirow{2}{*}{\textbf{Method}}} & \multicolumn{3}{c}{\textbf{In-distribution Datasets}} & & \multicolumn{2}{c}{\textbf{Out-of-distribution Datasets}} \\
    \cmidrule{2-4} 
    \cmidrule{6-7}
    \multicolumn{1}{c}{} & GSM8K & MATH500 & AIME24 & & MMLU-Pro & GPQA \\
    \midrule
    \multicolumn{7}{c}{R1-1.5B(Base Model)} \\
    \midrule
    AdapThink($\delta$=0.05) & 83.1 & 82.0 & 31.0 & & 35.77 & 32.89 \\
    JointThinking & \textbf{84.2} & \textbf{82.8} & \textbf{35.0} & & \textbf{45.21} & \textbf{39.78} \\
    \midrule
    \multicolumn{7}{c}{R1-7B(Base Model)} \\
    \midrule
    AdapThink($\delta$=0.05) & 91.0 & \textbf{92.0} & \textbf{55.6} & & 57.07 & 51.23 \\
    JointThinking & \textbf{91.1} & 87.8 & 51.3 & & \textbf{66.79} & \textbf{57.49} \\
    \bottomrule
    \end{tabular}}
    \caption{Comparison between the SOTA training-based reasoning enhancement method (\textit{AdapThink}) and our proposed ICL-based approach (\textit{JointThinking}) on both in-distribution (ID) and out-of-distribution (OOD) datasets. Two methods achieve comparable performance on ID datasets, while \textit{JointThinking} significantly outperforms \textit{AdapThink} on OOD datasets, demonstrating superior generalization ability. The better results are \textbf{bolded}.}
    \label{table:generalization}
\end{table*}

\section{Method}

\subsection{Is Few-Shot CoT Still Effective?}

\begin{figure}[t]
    \includegraphics[width=0.95\linewidth]{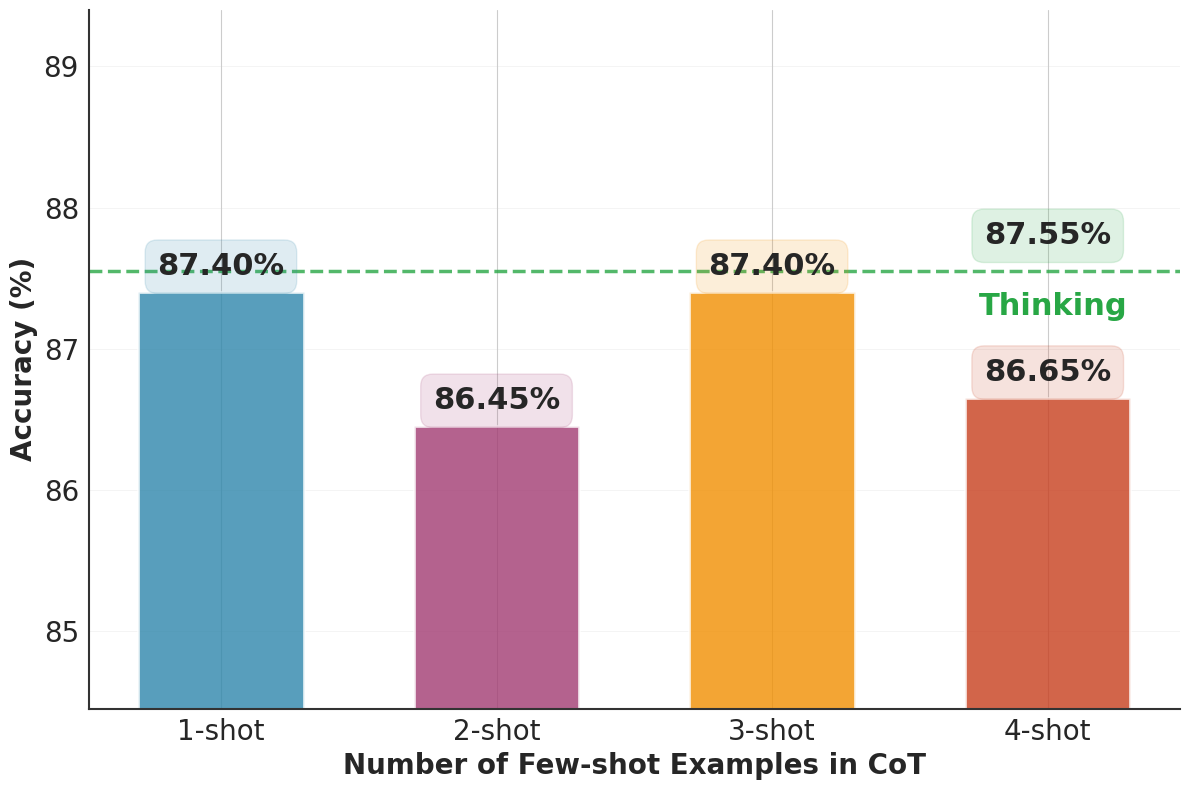}
    \caption{Effect of varying the number of few-shot CoT examples on MATH500 accuracy using R1-7B. Results are averaged over four runs. Few-shot CoT yields no improvement over direct 
    thinking, and the performance is unrelated to the number of few-shot examples.}
    \label{few-shot-n}
\end{figure}


Few-shot CoT prompting has demonstrated strong effectiveness for traditional LLMs, consistently improving performance across a variety of tasks \cite{wei2022chain}. However, does this approach remain effective for RLLMs? To explore this, we conduct a preliminary analysis. As shown in Figure 3, few-shot CoT offers no improvement over direct \emph{Thinking} (represented by the green horizontal dashed line), and its performance remains largely unaffected by the number of examples provided. This observation aligns with expectations, as current RL training pipelines for RLLMs are primarily designed to enhance internal reasoning capabilities, thereby reducing reliance on explicit instructional prompts such as few-shot demonstrations.




\subsection{Thinking with Nothinking Calibration}


Inspired by the complementary behavior in Figure \ref{complementary}, where regardless of problem difficulty, there consistently exist cases where the model fails under the \textit{Thinking} mode but succeeds under the \textit{Nothinking} mode, and vice versa, we propose a novel ICL framework to leverage this structured discrepancy, as shown in Figure~\ref{fig:framework}.



\subsubsection{Thinking Mode and Nothinking Mode}
\label{sec:two_modes_introduction}


Most RLLMs today follow a common generation paradigm, where the model first produces a reasoning chain enclosed by special tokens \texttt{\textless|beginning of thinking|\textgreater} and \texttt{\textless|end of thinking|\textgreater}, followed by the final solution steps and answer.


Let $M$ denote the reasoning model, with \textless{}BOT\textgreater{} and \textless{}EOT\textgreater{} denoting the special tokens for the beginning and end of the thinking process, respectively. Given an input question $q$, we define two different generation modes:

(i) \textit{Thinking} mode: The model generates a full reasoning trajectory beginning with \textless{}BOT\textgreater{}:
\begin{equation}
r_{\text{thinking}} = M(q + \langle\text{BOT}\rangle)
\end{equation}

(ii) \textit{NoThinking} mode: The model is prompted to skip the thinking process and directly generate the final output:
\begin{equation}
r_{\text{nothinking}} = M(q + \langle\text{BOT}\rangle + s_{\text{skip}} + \langle\text{EOT}\rangle)
\end{equation}
where $s_{\text{skip}}$ is set to ``Okay, I think I have finished thinking.'' following \cite{ma2025reasoning}. This prompt design effectively forces the model to skip the thinking stage and proceed directly to the final solution. Full prompt templates are provided in Appendix \ref{appendix:prompt_details}. 


\subsubsection{Second Thinking}

We introduce a consistency check mechanism to determine when a second round of thinking is needed. Let $A(r)$ be a function that extracts the final answer from a response $r$. The consistency check $\text{Con}(q)$ is defined as:

\begin{equation}
\text{Con}(q) = \begin{cases}
1, & \text{if } A(r_{\text{t}}) = A(r_{\text{n}}) \\
0, & \text{otherwise}
\end{cases}
\end{equation}
where $r_t = r_{\text{thinking}}$ and $r_n = r_{\text{nothinking}}$. Based on this check, the final answer $\text{R}(q)$ is selected as:
\begin{equation}
\text{R}(q) = \begin{cases} 
A(r_t), & \text{if } \text{Con}(q) = 1 \\
\text{ST}(q, A(r_t), A(r_n)), & \text{if } \text{Con}(q) = 0
\end{cases}
\end{equation}
and $\text{ST}$ denotes the second thinking processing. If the two answers are consistent, the response from the \emph{Thinking} mode is directly adopted as the final output. Otherwise, both answers are incorporated into a second-thinking prompt to elicit a refined answer. A brief overview of the prompt is provided below, with full details available in Appendix A.  

\begin{equation}
r_{\text{ST}} = M(q + \langle\text{BOT}\rangle + s_{\text{ST}})
\end{equation}
where $r_{\text{ST}}$ is the final answer from the second thinking process and $s_{\text{ST}}$ represents ``I think there are two candidate answers $r_{\text{thinking}}$ and $r_{\text{nothinking}}$ for this question, I need to verify them first. If both are wrong, I need to rethink.''.

Importantly, we find that increasing prompt complexity does not improve second-thinking performance. And the instruction should be placed immediately after \textless{}BOT\textgreater{}, marking the beginning of the reasoning process, rather than before \textless{}BOT\textgreater{}, as is typically done in traditional prompt engineering (an empirical analysis is discussed in Section \ref{subsec:current_limitation}). For clarity, we include the complete algorithm of the \emph{JointThinking} pipeline in Appendix \ref{section:jointthinking_algorithm}.

%% file: sections/experiments.tex
\section{Experiments}
\label{section:experiments}


\subsection{Setup}

\paragraph{Models} We conduct experiments using six models from the DeepSeek-R1-Distill-Qwen series: 1.5B, 7B, 14B, and 32B variants \cite{guo2025deepseek}, and from Qwen3 series: 8B and 14B variants \cite{yang2025qwen3}. All of them are SOTA open-source models and widely adopted in the community. All models are deployed on 2 \texttimes{} A800 80G GPUs.


\paragraph{Methods} We compare our proposed method, \textit{JointThinking}, with five no-training baselines. The \textit{Thinking} and \textit{Nothinking} modes are defined in Section~\ref{sec:two_modes_introduction}. For few-shot CoT \citep{ge2025innate} we choose the number of examples to 3 in terms of Figure \ref{few-shot-n}. Since \textit{JointThinking} involves at most three inference passes, majority voting is set to voting@3 for a fair comparison. Thinking twice is a simple two-step thinking process, where the second \textit{Thinking} is always triggered and uses the answer of the first \textit{Thinking} as context. Both few-shot CoT and majority voting are in \textit{Thinking} mode.


\paragraph{Datasets} Evaluation is conducted on four mathematical reasoning benchmarks: GSM8K \cite{cobbe2021training}, MATH500 \cite{lightman2023let}, AIME2024 \cite{aime2024}, and AMC2023 \cite{amc2023}, covering a wide range of difficulty levels from elementary school word problems to high school competition-level mathematics. All models are configured with a 16K context window to support the generation of long reasoning chains. Following the official guidance, decoding parameters are set as follows: for the \textit{Thinking} mode, temperature is 0.6 and top-p is 0.95; for the \textit{Nothinking} mode, temperature is 0.7 and top-p is 0.8.


\subsection{Main Results}


As demonstrated in Table~\ref{table:main results}, the proposed \textit{JointThinking} method consistently outperforms all no-training methods across all models. It achieves average performance gains ranging from 0.77 to 2.94 over the second-best baseline and from 2.20 to 6.21 compared to the standard single \textit{Thinking} mode. The results also indicate that the \textit{Thinking} mode generally outperforms the \textit{Nothinking} mode, with the performance gap becoming more pronounced on challenging datasets such as AIME2024, where the difference reaches up to 35.42 on Qwen3-14B. However, on easier datasets, the performance of the two modes tends to converge (e.g., 92.80 vs. 92.72 for R1-32B on GSM8K), and in some cases, \textit{Nothinking} even slightly outperforms \textit{Thinking} (e.g., 79.00 vs. 78.20 for R1-1.5B on MATH500).

While few-shot CoT remains one of the most commonly used ICL strategies for traditional LLMs, it demonstrates limited effectiveness on RLLMs. Nonetheless, it yields modest gains on few benchmarks for larger models such as Qwen3-14B on GSM8K and R1-32B on MATH500. In contrast, majority voting and thinking twice remain strong no-training methods for RLLMs. The former exhibits relatively superior performance on smaller model sizes, whereas the latter tends to be more effective on larger models. However, thinking twice shows instability, occasionally performing on par with or even worse than single \textit{Thinking} (e.g., 94.77 vs. 94.77 for Qwen3-14B on GSM8K, and 60.00 vs. 63.75 for R1-14B on AIME24). Our proposed \textit{JointThinking} consistently outperforms both baselines, demonstrating that simply increasing the number of reasoning attempts can't improve performance stably and efficiently. We show one detailed case study of our method in Appendix \ref{appendix:case_study}.

\subsection{Generalizability}

\begin{table}[t]
    \centering
    \scalebox{0.85}{
    \begin{tabular}{lccc}
    \toprule
    \multicolumn{1}{c}{\multirow{3}{*}{\textbf{Dataset}}} & \multicolumn{2}{c}{\textbf{Error Rate (ER)}} & \multicolumn{1}{c}{\multirow{3}{*}{\textbf{$\Delta$ER}}}\\
    \cmidrule{2-3}
    \multicolumn{1}{c}{} & Thinking with & Thinking with & \multicolumn{1}{c}{} \\
    \multicolumn{1}{c}{} & Thinking & Nothinking & \multicolumn{1}{c}{} \\
    \midrule
    GSM8K & 4.90\% & 4.06\% & 0.84\% \\
    MATH500 & 5.23\% & 3.68\% & 1.55\% \\
    AIME24 & 1.67\% & 1.25\% & 0.42\% \\
    AMC23 & 0.63\% & 0.00\% & 0.63\% \\
    \bottomrule
    \end{tabular}}
    \caption{Error rate across different datasets based on R1-7B. For each dataset, we sample 8 outputs for computation. More details are provided in Appendix \ref{appendix:computation_details_of_error_ratio}.}
    \label{table:calibration}
\end{table}


In-context learning (ICL) provides inherent generalization advantages over training-based paradigms, supported by both theoretical analysis \cite{hosseini-etal-2022-compositional,zhang2025understanding} and empirical evidence \cite{li-etal-2024-context,lampinen2025generalization}. In this section, we compare our proposed method with \textit{AdaptThink} \cite{wan2025adapthink}, a SOTA training-based reasoning enhancement approach. \textit{AdaptThink} introduces a reinforcement learning strategy that enables RLLMs to dynamically select the optimal thinking mode according to problem difficulty, thereby effectively mitigating overthinking. We evaluate both methods on in-distribution (ID) and out-of-distribution (OOD) scenarios.


For ID evaluation, we follow the \textit{AdaptThink} paper and use GSM8K, MATH500, and AIME24 datasets. For OOD evaluation, we consider two challenging benchmarks: MMLU-Pro \cite{wang2024mmlu} and GPQA \cite{rein2024gpqa}. MMLU-Pro is a professional-level subset of MMLU targeting advanced subject-specific reasoning, while GPQA is a challenging, expert-validated benchmark requiring multi-step reasoning in scientific domains.


As illustrated in Table~\ref{table:generalization}, \textit{AdaptThink} and \textit{JointThinking} achieve comparable performance on ID datasets, with \textit{JointThinking} consistently outperforming \textit{AdaptThink} across all three benchmarks for R1-1.5B. The most notable difference arises in the OOD setting. On both MMLU-Pro and GPQA, \textit{JointThinking} significantly outperforms \textit{AdaptThink}, improving from 35.77 to 45.21 and from 32.89 to 39.78 on R1-1.5B, and from 57.07 to 66.79 and from 51.23 to 57.49 on R1-7B, respectively. These results highlight the strong generalization capability of \textit{JointThinking}, enabling robust performance across diverse unseen domains. 

\subsection{Calibration Analysis}
In this subsection, we provide a systematic explanation and analysis of the calibration mechanism underlying our proposed method. 
As illustrated in Figure~\ref{complementary}, we observe a complementary relationship between the \textit{Thinking} and \textit{Nothinking} modes: one often succeeds where the other fails. We hypothesize this arises from fundamental structural differences between the two reasoning modes. The \textit{Nothinking} mode relies more on intuition, rapidly recalling answers from training patterns but sometimes overlooking key details. Conversely, the \textit{Thinking} mode emphasizes deliberate reasoning through iterative exploration, backtracking, and verification, though it may overthink problems. 

To empirically verify this structural thinking divergence, we evaluate the effectiveness of cross-mode calibration by using both \textit{Thinking} and \textit{Nothinking} modes to calibrate outputs generated by the \textit{Thinking} mode. We introduce a metric called \textbf{Error Rate (ER)} to measure calibration performance. Let $\mathcal{D}$ denote a dataset with ground-truth labels, and let $\mathcal{C} \subseteq \mathcal{D}$ represent the subset of examples for which both modes produce consistent answers. The Error Rate is defined as:

\begin{equation}
\text{ER} = \frac{|\{q \in \mathcal{C} : A(r_{\text{thinking}}) \neq \text{GT}(q)\}|}{|\mathcal{C}|}
\label{eq:er_avg}
\end{equation}
where $\text{GT}(q)$ denotes the ground-truth answer to question $q$, and $A(r)$ extracts the final answer from a response $r$. Inconsistent answers naturally signal uncertainty and can be escalated to a second round of reasoning using the same model or even more capable RLLMs for refinement. However, when an incorrect answer is mistakenly deemed reliable due to agreement between modes, it may be finalized prematurely, preventing further correction. Therefore, a lower ER reflects more effective calibration, indicating that the process effectively preserves correct answers while reducing errors.

\begin{figure}[t]
    \centering
    \includegraphics[width=0.96\linewidth]{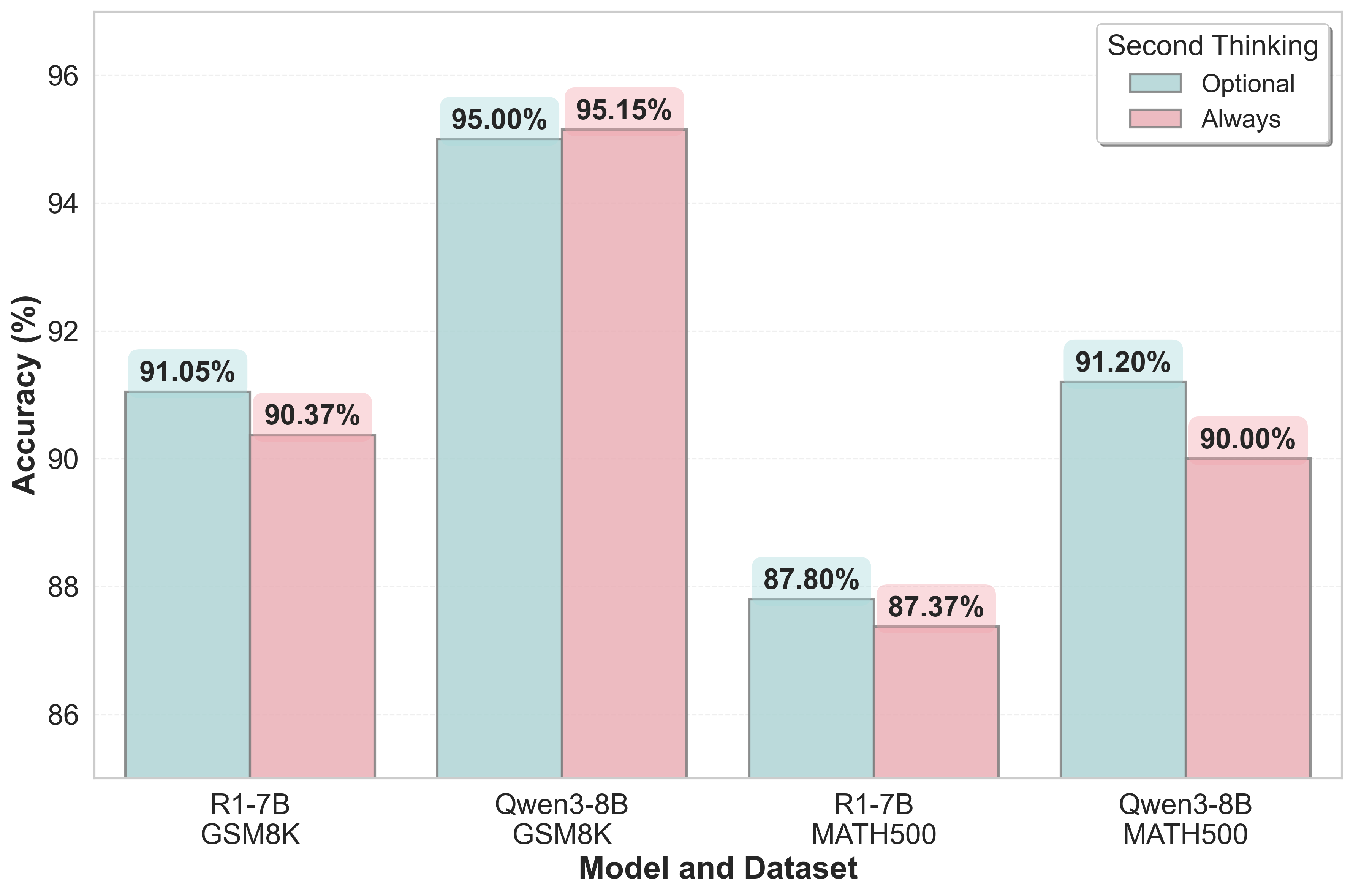}
    \caption{Performance comparison between optional and always second thinking. The optional trigger setting yields better results with less computational overhead.}
    \label{fig:consistency_check_comparison_small}
\end{figure}

As shown in Table~\ref{table:calibration}, we sample 8 outputs per example using R1-7B across all datasets to reduce variance and report the average results using equation~\eqref{eq:er_avg}. Across all benchmarks, calibrating \textit{Thinking} outputs with the \textit{Nothinking} mode consistently outperforms calibration with the same mode. Notably, we observe absolute reductions in error rates ranging from 0.42\% to 1.55\%, underscoring the benefits of calibrating across structurally different reasoning processes. These results support our hypothesis that the structural difference between reasoning modes enables more effective calibration.

%% file: sections/analysis.tex
\section{Further Analysis and Discussion}
\subsection{The Necessity of Consistency Check}

\begin{figure}[t]
    \centering
    \includegraphics[width=0.95\linewidth]{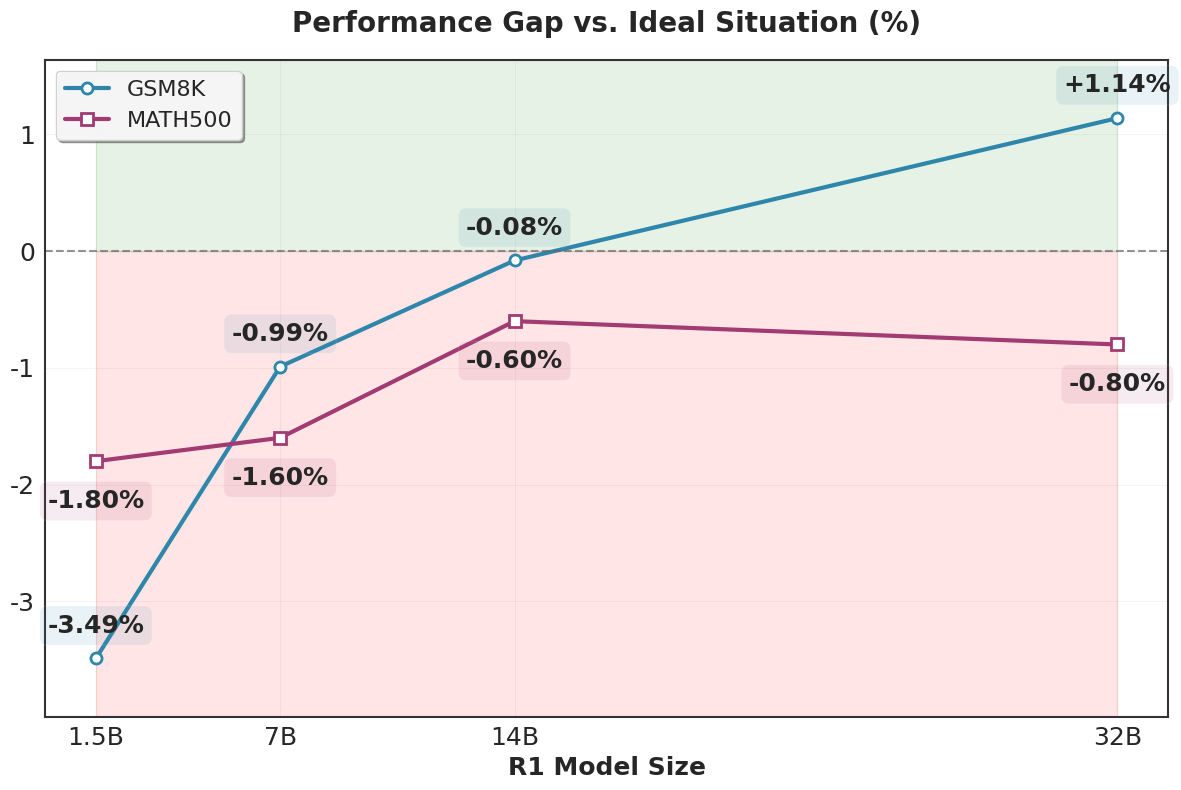}
    \caption{Scaling trend of second-thinking performance. The reduction in performance gap from the ideal situation indicates the strong scalability of \textit{jointthinking}.}
    \label{fig:second thinking trend}
\end{figure}

The primary advantage of the consistency check mechanism in the second stage of our method lies in efficiency, as it is only applied to the most uncertain cases, keeping additional inference latency minimal. For example, on GSM8K with R1-32B, second thinking is required in just 6\% of cases. More calculation details can be found in Appendix \ref{appendix:consistency_check_analysis}. Furthermore, as shown in Figure~\ref{fig:consistency_check_comparison_small}, selectively triggering second thinking via consistency check can even outperform always applying it, since repeated thinking risks turning correct answers into incorrect ones. This result highlights that naively increasing reasoning iterations may degrade model performance, underscoring the effectiveness of our pipeline's design.

\subsection{Scaling Trend in Second Thinking}

The previous analysis primarily focuses on the overall performance of \emph{JointThinking}. In this section, we shift our attention to the second thinking stage and investigate how its effectiveness evolves as model size increases. In our approach, when the \textit{Thinking} and \textit{Nothinking} modes generate inconsistent answers, a new prompt is constructed and guides the model for a second round of inference using the \textit{Thinking} mode. This procedure is referred to as the second thinking. Within these inconsistent cases, we identify two different scenarios:

\textbf{Scenario 1}: Both modes produce incorrect answers which are different each other.

\textbf{Scenario 2}: One mode produces a correct answer while the other does not.

Ideally, in Scenario 2, the second thinking stage should consistently recover the correct answer. However, due to structural misalignment between reasoning modes or inherent randomness in decoding, this ideal is not always achieved. Figure~\ref{fig:second thinking trend} illustrates the performance gap between actual second thinking results and the ideal upper bound (i.e., perfect recovery in Scenario 2) on GSM8K and MATH500, across models of increasing size. Interestingly, we observe a clear upward trend in second thinking performance as model size increases. Notably, the 32B model even surpasses the ideal baseline by 1.14\% on GSM8K, suggesting that the RLLM can arrive at the correct answer through second thinking even when all candidate answers are incorrect. This result highlights the strong scalability of our approach, which we attribute to enhanced reflective capabilities of larger models.



\subsection{Current Limitations and Future Directions}
\label{subsec:current_limitation}

\begin{table}[t]
    \centering
    \scalebox{0.85}{
    \renewcommand{\arraystretch}{0.9}
    \begin{tabular}{lccc}
    \toprule
    \multicolumn{1}{c}{\multirow{3}{*}{\textbf{Dataset}}} & \multicolumn{3}{c}{\textbf{Methods}} \\
    \cmidrule{2-4}
    \multicolumn{1}{c}{} & Think & JointThinking & JointThinking \\
    \multicolumn{1}{c}{} & -ing & before \textless{}think\textgreater{} & after \textless{}think\textgreater{} \\
    \midrule
    \multicolumn{4}{c}{R1-7B} \\
    \midrule
    GSM8K & 87.79 & 89.61 & 91.05 \\
    MATH500 & 87.60 & 87.20 & 87.80 \\
    AIME24 & 47.50 & 50.00 & 51.25 \\
    AMC23 & 86.88 & 87.50 & 88.44 \\
    \midrule
    \multicolumn{4}{c}{Qwen3-8B} \\
    \midrule
    GSM8K & 94.69 & 95.00 & 95.00 \\
    MATH500 & 87.80 & 90.00 & 91.20 \\
    AIME24 & 59.17 & 62.08 & 71.25 \\
    AMC23 & 86.56 & 90.31 & 92.81 \\
    \bottomrule
    \end{tabular}}
    \caption{Effect of prompt placement in the second thinking stage of \textit{JointThinking}. More experiments across different models can be found in Appendix \ref{appendix:addition_results}.}
    \label{table:prompt position}
\end{table}

\vspace{-0.2em}


In the final part of this paper, we summarize several key findings, to discuss current limitations and suggest directions for future research. The success of LLMs is largely attributed to their strong ICL capabilities. However, this strength transfers less effectively to RLLMs. This is perhaps expected, as most RL algorithms for training RLLMs focus on solving complex mathematical problems and simply generating long reasoning chains. Such training often employs simplified prompts that lack diversity and rich information, which may inadvertently reduce the model's sensitivity to external guidance.

Our findings in Table~\ref{table:main results} emphasize this observation, where few-shot CoT demonstrates limited effectiveness on RLLMs compared to expectations. To further investigate this limitation, we conduct an ablation study by explicitly changing the position of the prompt relative to the reasoning token in the second thinking stage. Specifically, we compare two configurations: (1) placing the prompt, consisting of the original question and two inconsistent candidate answers, \textbf{before} the \texttt{\textless think\textgreater} token, treating it as a general instruction, and (2) placing the prompt \textbf{after} this special token, integrating it as part of the model's reasoning process. The results reported in Table~\ref{table:prompt position} show that the first setting, resembling standard prompting, provides limited benefit and even degrades performance on MATH500 with R1-7B. In contrast, the second setting consistently leads to improved performance. These findings reveal a fundamental shortcoming of current RLLMs: insufficient instruction-following capabilities hinder the effective use of ICL.



Building on these insights, we suggest two directions for future research: (1) On the training side, RL strategies should be designed not only to improve reasoning performance but also to preserve or even strengthen the model's ability to follow instructions; (2) On the inference side, future work may explore how to better leverage the structural differences between reasoning modes and integrate them more effectively for enhanced reasoning. We hope our findings inspire further exploration into improving the ICL capabilities of RLLMs.

%% file: sections/conclusion.tex
\section{Conclusion}

\vspace{-0.2em}


In this work, we introduce \textit{Thinking with Nothinking Calibration (JointThinking)}, a new ICL paradigm for RLLMs that leverages the structural difference between \textit{Thinking} and \textit{Nothinking} modes. 
Experimental results demonstrate that \emph{JointThinking} significantly outperforms existing no-training approaches, and 
exhibits stronger generalization 
compared with training-based reasoning enhancement. We further provide a systematic analysis of the calibration mechanism, underscoring the importance of incorporating different reasoning modes and consistency check mechanism. Finally, we investigate how model scaling influences second-thinking performance, highlighting the strong scalability of our method.

%% file: sections/limitations.tex
\section{Limitations}
While our approach demonstrates strong effectiveness, several limitations should be acknowledged. First, due to computational constraints for RLLM inference with 16K context window, our experiments are limited to model size up to 32B. The experiment results demonstrate that our proposed method works across different models and sizes, and the scaling trend of the second thinking also suggests its strong scalability. Second, we focus exclusively on \textit{Thinking} and \textit{Nothinking} modes, without exploring broader reasoning mode combinations. However, this limitation largely stems from the constrained set of reasoning modes currently supported by existing RLLMs rather than our framework design. Third, our consistency check mechanism relies on exact or near-exact answer matching to trigger the second round of thinking. It works well for mathematical reasoning benchmarks in our experiments,  while for more open-ended problems the semantic similarity consistency check could be considered.  

%% file: sections/appendix.tex
\appendix

\section{Prompts}
\label{appendix:prompt_details}

\begin{table*}[ht]
    \centering
    \scalebox{0.88}{
    \begin{tabular}{cc}
    \toprule
    \textbf{Method} & \textbf{Prompt Template} \\
    \midrule
    \multicolumn{1}{c}{\multirow{2}{*}{Thinking}} & Problem: \{problem\} \textbackslash n Please put your final answer in the format boxed\{\} \\
    \multicolumn{1}{c}{} & \textless{}\textbar Assistant\textbar\textgreater{}\textbackslash n Answer:\textless{}think\textgreater{}\textbackslash n \\
    \midrule
    \multicolumn{1}{c}{\multirow{2}{*}{Nothinking}} & Problem: \{problem\} \textbackslash n Please put your final answer in the format boxed\{\} \\
    \multicolumn{1}{c}{} & \textless{}\textbar Assistant\textbar\textgreater{}\textbackslash n Answer:\textless{}think\textgreater{}\textbackslash n Okay I have finished thinking.\textbackslash n\textless{}/think\textgreater{}\textbackslash n\\
    \midrule
    \multicolumn{1}{c}{\multirow{3}{*}{Few-shot}} & Problem: \{example problem\}\textbackslash n Answer: \{example answer\}\textbackslash n\textbackslash n ... (more few-shot cot examples)\\
    \multicolumn{1}{c}{} & Problem: \{problem\} \textbackslash n Please put your final answer in the format boxed\{\} \\
    \multicolumn{1}{c}{} & \textless{}\textbar Assistant\textbar\textgreater{}\textbackslash n Answer:\textless{}think\textgreater{}\textbackslash n \\
    \midrule
    \multicolumn{1}{c}{\multirow{5}{*}{Thinking Twice}} & Problem: \{problem\} \textbackslash n Please put your final answer in the format boxed\{\} \\
    \multicolumn{1}{c}{} & \textless{}\textbar Assistant\textbar\textgreater{}\textbackslash n Answer:\textless{}think\textgreater{}\textbackslash n I think there is one candidate answer \\
    \multicolumn{1}{c}{} & \{thinking\_answer\} for this question. \\
    \multicolumn{1}{c}{} & I need to first verify it.\\
    \multicolumn{1}{c}{} & If it is wrong, I need to rethink step by step and avoid making the same mistake.\\
    \midrule
    \multicolumn{1}{c}{\multirow{5}{*}{JointThinking}} & Problem: \{problem\} \textbackslash n Please put your final answer in the format boxed\{\} \\
    \multicolumn{1}{c}{} & \textless{}\textbar Assistant\textbar\textgreater{}\textbackslash n Answer:\textless{}think\textgreater{}\textbackslash n I think there are two candidate answers \\
    \multicolumn{1}{c}{} & \{thinking\_answer\} and \{nothinking\_answer\} for this question. \\
    \multicolumn{1}{c}{} & One of them is correct or both are wrong. I need to first verify them.\\
    \multicolumn{1}{c}{} & If both are wrong, I need to rethink step by step and avoid making the same mistake.\\
    \bottomrule
    \end{tabular}}
    \caption{Methods and their corresponding prompt templates for main results.}
    \label{table:additonal_prompt_formats}
\end{table*}

Table \ref{table:additonal_prompt_formats} presents detailed prompt templates for different methods in Table \ref{table:main results}. In the section \ref{sec:two_modes_introduction} when we introduce our method we use special tokens \texttt{\textless|beginning of thinking|\textgreater} and \texttt{\textless|end of thinking|\textgreater} to explain our prompt engineering in general. In our experiments, we employ the R1-series and Qwen3-series models, with specific thinking special tokens \texttt{\textless{}think\textgreater{}} and \texttt{\textless{}/think\textgreater{}} instead. In particular, for the Qwen3 it natively supports switching between the thinking and nothinking modes by adding special tokens \texttt{\textbackslash think} or \texttt{\textbackslash nothink} after the prompt, so we follow this setting in Thinking, Nothinking and Few-shot scenarios.
Notably, in the \textit{JointThinking} pipeline, it will first use general thinking mode and nothinking mode to generate answers with their corresponding prompt templates. Only in the second thinking the \textit{JointThinking} prompt structure shown in the last row of the table will be used. Additionally, all instructions are provided through the user prompt, without using system prompt.

\section{Algorithm Description}
\label{section:jointthinking_algorithm}
The complete pipeline of our method \textit{JointThinking} is summarized in Algorithm \ref{alg:thinking_nothinking}.

\begin{algorithm*}[htb]
\caption{\textit{Thinking with NoThinking Calibration (JointThinking)}}
\label{alg:thinking_nothinking}
\begin{algorithmic}[1]
\REQUIRE Question $q$, Model $M$
\ENSURE Final answer $a_{\text{final}}$

\STATE Generate thinking response: $r_{\text{thinking}} \leftarrow M(q + \langle\text{BOT}\rangle)$
\STATE Generate nothinking response: $r_{\text{nothinking}} \leftarrow M(q + \langle\text{BOT}\rangle + s_{\text{skip}} + \langle\text{EOT}\rangle)$
\STATE Extract answers: $a_1 \leftarrow A(r_{\text{thinking}})$, $a_2 \leftarrow A(r_{\text{nothinking}})$

\IF{$a_1 = a_2$}
    \STATE $a_{\text{final}} \leftarrow a_1$
\ELSE
    \STATE Construct second thinking prompt: $p_{\text{second\_thinking}} \leftarrow \text{Prompt}_{\text{second\_thinking}}(q, r_{\text{thinking}}, r_{\text{nothinking}})$
    \STATE Generate final response: $r_{\text{final}} \leftarrow M(p_{\text{second\_thinking}})$
    \STATE $a_{\text{final}} \leftarrow A(r_{\text{final}})$
\ENDIF

\RETURN $a_{\text{final}}$
\end{algorithmic}
\end{algorithm*}

\section{Case Study}
\label{appendix:case_study}

To further illustrate our proposed method, we present a detailed case study based on R1-32B model. We select one problem from the AMC23 datasets that are wrong in single thinking mode, but are solved correctly by \textit{JointThinking}.

\definecolor{thinking_color}{RGB}{194, 213, 247}

\begin{figure*}[h]
\centering

\begin{tcolorbox}[
  enhanced,
  breakable,
  fonttitle = \small\bfseries, 
  title= Pure Question from AMC23,
  colframe=blue!30!white,   
  colback=blue!5!white,     
  boxrule=1pt,
  boxsep=0pt,
  left=5pt,
  right=5pt,
  fontupper=\footnotesize,
  halign title = flush center
]
Carlos went to a sports store to buy running shoes. Running shoes were on sale, with prices reduced by $20\%$ on every pair of shoes. Carlos also knew that he had to pay a $7.5\%$ sales tax on the discounted price. He had \$43. What is the original (before discount) price of the most expensive shoes he could afford to buy?
\end{tcolorbox}
\caption{One example question from AMC23 dataset.}
\label{fig:amc23-question}
\end{figure*}

Under the thinking mode, the model arrives at an incorrect final answer. This error stems from a flawed reasoning process: although the model correctly removes the 7.5\% sales tax by dividing the total budget by 1.075, it then mistakenly treats the resulting value as the discounted price before applying the 20\% reduction. In reality, the value obtained after removing the tax already reflects both the 20\% discount and the tax. As a result, when the model attempts to "reverse" the discount by dividing again by 0.8, it overcompensates and produces an underestimated final answer (i.e., \$49.94 instead of the correct \$50.00). Notably, the question does not trigger any reflection or revision during the model’s thinking phase. Although the subsequent computational steps after thinking processing are mathematically sound, the incorrect answer of thinking part still leads the model to the wrong conclusion.

\begin{figure*}[h]
\centering

\begin{tcolorbox}[
  enhanced,
  breakable,
  fonttitle = \small\bfseries, 
  title= Thinking Mode,
  colframe=blue!30!white,
  colback=blue!5!white,
  boxrule=1pt,
  boxsep=0pt,
  left=5pt,
  right=5pt,
  fontupper=\footnotesize,
  halign title = flush center
]

\colorbox{thinking_color}{\textless{}think\textgreater{}}

First, let's determine the maximum amount Carlos can spend after accounting for the sales tax. He has \$43, and the sales tax rate is $7.5\%$. To find the amount before tax, we divide \$43 by $1.075$, which gives approximately \$39.95.

\vspace{0.5em}

Next, we need to find the original price of the shoes before the $20\%$ discount. Since the discounted price is \$39.95, we divide this by $0.80$ to reverse the $20\%$ reduction. This calculation results in approximately \$49.94.

\vspace{0.5em}

Therefore, the original price of the most expensive shoes Carlos could afford is approximately \$49.94.

\colorbox{thinking_color}{\textless{}/think\textgreater{}}

\vspace{0.5em}

To determine the original price of the most expensive shoes Carlos could afford, let's follow these steps:

\begin{enumerate}
  \item Understand the Problem:
  \begin{itemize}
    \item Sale Discount: $20\%$ off the original price.
    \item Sales Tax: $7.5\%$ on the discounted price.
    \item Carlos's Budget: \$43.
  \end{itemize}
  
  \item Let’s Define Variables:
  \begin{itemize}
    \item Let \( P \) be the original price of the shoes.
    \item The discounted price after $20\%$ off is \( 0.80P \).
    \item The final price after adding $7.5\%$ tax is \( 0.80P \times 1.075 \).
  \end{itemize}
  
  \item Set Up the Equation:
  \[
  0.80P \times 1.075 \leq 43
  \]

  \item Solve for \( P \):
  \[
  0.80P \times 1.075 = 43
  \]
  \[
  0.86P = 43
  \]
  \[
  P = \frac{43}{0.86} \approx 49.94
  \]

  \item Conclusion:\\
  The original price of the most expensive shoes Carlos could afford is approximately \$49.94.
  \[
  \boxed{49.94}
  \]
\end{enumerate}
\end{tcolorbox}
\caption{The output of R1-32B in Thinking mode for selected question.}
\label{fig:thinking-amc23-question}
\end{figure*}

In the nothinking mode, the model provides a clean and well-structured step-by-step solution with correct answer.

\begin{figure*}
\centering 

\begin{tcolorbox}[
  enhanced,
  breakable,
  fonttitle = \small\bfseries, 
  title= Nothinking Mode,
  colframe=blue!30!white,
  colback=blue!5!white,
  boxrule=1pt,
  boxsep=0pt,
  left=5pt,
  right=5pt,
  fontupper=\footnotesize,
  halign title = flush center
]

To determine the original price of the most expensive shoes Carlos could afford, we'll follow these steps:

\begin{enumerate}
  \item Understand the Discounts and Taxes:
  \begin{itemize}
    \item Discount: The shoes are discounted by $20\%$. This means Carlos pays $80\%$ of the original price.
    \item Sales Tax: After the discount, there's a $7.5\%$ sales tax applied to the discounted price.
  \end{itemize}

  \item Let \( P \) be the original price of the shoes.

  \item Calculate the Discounted Price:
  \[
  \text{Discounted Price} = P \times (1 - 0.20) = 0.80P
  \]

  \item Calculate the Price After Tax:
  \[
  Final Price = Discounted Price \times (1 + 0.075) = 0.80P \times 1.075 = 0.86P
  \]

  \item Set Up the Inequality Based on Carlos's Budget:
  \[
  0.86P \leq 43
  \]

  \item Solve for \( P \):
  \[
  P \leq \frac{43}{0.86} = 50
  \]
\end{enumerate}

Conclusion: The original price of the most expensive shoes Carlos could afford is \( \boxed{50} \) dollars.

\end{tcolorbox}
\caption{The output of R1-32B in Nothinking mode for selected question.}
\end{figure*}

In the \textit{JointThinking}, the reasoning model starts thinking with two candidate answers. We observe that the model initially approaches the problem in a relatively independent manner, treating the task care. The presence of two plausible candidate answers in the prompt introduces uncertainty, prompting the model to engage in more cautious and deliberate reasoning. Interestingly, after arriving at the correct answer, the model proceeds to critically examine the alternative option, evaluating its plausibility and attempting to justify or refute it (the relevant sentences are underlined with \textcolor{red}{red} color). This process triggers a distinctive reasoning pattern characterized by iterative reflection and verification. Ultimately, the model selects the correct answer and provides a possible explanation for why the model is able to get the other incorrect candidate answer.

\onecolumn
\begin{tcolorbox}[
  enhanced,
  breakable,
  fonttitle = \small\bfseries, 
  title= JointThinking,
  colframe=blue!30!white,
  colback=blue!5!white,
  boxrule=1pt,
  boxsep=0pt,
  left=5pt,
  right=5pt,
  fontupper=\footnotesize,
  halign title = flush center
]

\colorbox{thinking_color}{\textless{}think\textgreater{}}

So, let's start.

\vspace{0.5em}

First, the problem says Carlos went to a sports store to buy running shoes. The running shoes were on sale with a 20\% discount on every pair. Then, Carlos has to pay a 7.5\% sales tax on the discounted price. He had \$43. So, the question is, what is the original price of the most expensive shoes he could afford?

\vspace{0.5em}

Alright, so I need to find the original price, let's call it $P$. Then, after a 20\% discount, the price becomes $P$ minus 20\% of $P$, which is $0.8P$. Then, he has to pay 7.5\% tax on that discounted price, so the total price he has to pay is $0.8P$ multiplied by $1.075$.

\vspace{0.5em}

So, the equation is $0.8P \times 1.075 = $ total price he has to pay. And this total price should be less than or equal to \$43 because he can't spend more than that.

\vspace{0.5em}

So, $0.8 \times 1.075$ is equal to $0.86$. So, $0.86P \leq 43$.

To find $P$, divide both sides by $0.86$. So, $P \leq 43 / 0.86$.

Calculating $43$ divided by $0.86$. Hmm, $0.86$ times $50$ is $43$, because $0.86 \times 50$ is $43$. So, $P \leq 50$. So, the original price must be less than or equal to \$50.

\vspace{0.5em}

But wait, the question is asking for the original price of the most expensive shoes he could afford. So, if the original price is \$50, he can buy it because the total is exactly \$43. So, the answer should be \$50.

\vspace{0.5em}

\textcolor{red}{But why is 49.94 another candidate?} Maybe someone thought that the total should be less than \$43, so they might have rounded down? Let me see.

\vspace{0.5em}

If the original price is \$49.94, then 20\% off is $49.94 \times 0.2 = 9.988$, so the discounted price is $49.94 - 9.988 = 39.952$. Then, 7.5\% tax on $39.952$ is $39.952 \times 1.075$. Let me calculate that.

\vspace{0.5em}

First, $39.952 \times 1.075$. Let's compute $39.952 \times 1 = 39.952$. Then, $39.952 \times 0.075$. Let me calculate $39.952 \times 0.07 = 2.79664$, and $39.952 \times 0.005 = 0.19976$. Adding those together, $2.79664 + 0.19976 = 2.9964$. So, total tax is $2.9964$. Adding to the discounted price: $39.952 + 2.9964 = 42.9484$, which is approximately \$42.95. So, that's under \$43.

\vspace{0.5em}

Therefore, if the original price is \$49.94, he would spend about \$42.95, which is under his \$43. So, why is \$50 the correct answer? Because if the original price is \$50, after discount and tax, it's exactly \$43, which he can afford. \textcolor{red}{So, is \$50 the correct answer? Or is \$49.94 the maximum he can spend without exceeding his budget?}

\vspace{0.5em}

Wait, the problem says he had \$43. So, he can spend up to \$43. So, if the total cost is exactly \$43, he can buy it. So, \$50 is okay because the total is exactly \$43. So, \$50 is the maximum original price he can afford.

\vspace{0.5em}

......

\vspace{0.5em}

So, why did I get confused earlier? Because I incorrectly subtracted $10.002$ from $50.01$ as $39.999$, but it's actually $40.008$. So, that was my mistake.

Therefore, the correct answer is \$50.

\vspace{0.5em}

But wait, let's confirm with $P = 50$.

Discount: $50 \times 0.2 = 10$, so discounted price is $40$.

Tax: $40 \times 1.075 = 43$.

So, total is exactly \$43, which he can afford.

If $P$ is just a little over $50$, say $50.01$, the total is just over $43$, which he can't afford.

Therefore, the maximum original price is \$50.

\vspace{0.5em}

But wait, in the initial calculation, I thought the total was $0.86P$, and $43 / 0.86 = 50$. So, that's consistent.

Therefore, the correct answer is \$50.

\vspace{0.5em}

\textcolor{red}{So, why was $49.94$ a candidate? Maybe someone thought that the total should be strictly less than \$43, so they might have rounded down, but in reality, since \$50 gives exactly \$43, which he can afford, the answer is \$50.}

\vspace{0.5em}

Therefore, the original price is \$50.

\colorbox{thinking_color}{\textless{}/think\textgreater{}}

\vspace{0.5em}

Carlos went to a sports store to buy running shoes that were on sale with a 20\% discount. After the discount, a 7.5\% sales tax was applied. He had \$43 to spend. To find the original price ($P$) of the most expensive shoes he could afford:

1.Discount Calculation: 20\% off means the price becomes 80\% of the original, so $0.8P$.

2.Tax Calculation: 7.5\% tax on the discounted price: $0.8P \times 1.075 = 0.86P$.

3. Total Cost: $0.86P \leq 43$.

4. Solve for P: $P \leq \frac{43}{0.86} = 50$.

\vspace{0.5em}

Verification:

1.For $P = 50$: Discounted price is $50 \times 0.8 = 40$. Tax adds $40 \times 0.075 = 3$, totaling $43$, which Carlos can afford.

2.For $P = 50.01$: The total exceeds \$43, so it's unaffordable.

\vspace{0.5em}

Thus, the most expensive shoes Carlos can afford originally cost \( \boxed{50} \) dollars.

\end{tcolorbox}
\twocolumn

\section{Experiment Details}
\label{appendix:details}

In this section, we will provide more supplementary computation details for tables used in the main text of the paper. 

\subsection{The Computation Details of Error Rate}
\label{appendix:computation_details_of_error_ratio}

In the Table \ref{table:additonal_error_ratio_details}, we illustrate the computation details of error rate for Table \ref{table:calibration}. Error rate denotes the proportion of incorrect answers among consistent answers in our method, referring to the result of dividing the number of answers in Both Correct by the number of answers in Both Incorrect \& Equiv in terms of table content. In order to ensure the validity of the conclusion, we repeat the experiment 8 times on R1-7B and take the average score as the final experimental result. The performance of thinking with nothinking calibration consistently outperforms thinking with thinking calibration for all datasets, showing that calibration capability from different modes is better than the same modes.

\subsection{The Consistency-checking Analysis of JointThinking in the Main Results}
\label{appendix:consistency_check_analysis}

\begin{table*}[h]
    \centering
    \scalebox{0.9}{
    \begin{tabular}{lccccc}
    \toprule
    \multicolumn{1}{c}{\multirow{2}{*}{\textbf{Model}}} & \multicolumn{1}{c}{\multirow{2}{*}{\textbf{Thinking\&NoThinking}}} & \multicolumn{4}{c}{\textbf{Datasets}} \\
    \cmidrule{3-6}
    \multicolumn{1}{c}{} & \multicolumn{1}{c}{(The First Stage)} & GSM8K(1319) & MATH500(500) & AIME24(30) & AMC23(40) \\
    \midrule
    \multicolumn{1}{c}{\multirow{5}{*}{R1-1.5B}} & Both Correct & 950 & 363 & 5 & 25 \\
    \multicolumn{1}{c}{} & Both Incorrect & 162 & 77 & 18 & 7 \\
    \multicolumn{1}{c}{} & Both Incorrect \& Equiv & 73 & 21 & 1 & 1 \\
    \multicolumn{1}{c}{} & Only Thinking Correct & 129 & 28 & 5 & 6 \\
    \multicolumn{1}{c}{} & Only NoThinking Correct & 78 & 32 & 2 & 2 \\
    \midrule
    \multicolumn{1}{c}{\multirow{5}{*}{R1-7B}} & Both Correct & 1069 & 361 & 6 & 25 \\
    \multicolumn{1}{c}{} & Both Incorrect & 105 & 53 & 15 & 4 \\
    \multicolumn{1}{c}{} & Both Incorrect \& Equiv & 51 & 18 & 0 & 0 \\
    \multicolumn{1}{c}{} & Only Thinking Correct & 89 & 77 & 9 & 9 \\
    \multicolumn{1}{c}{} & Only NoThinking Correct & 56 & 9 & 0 & 2 \\
    \midrule
    \multicolumn{1}{c}{\multirow{5}{*}{R1-14B}} & Both Correct & 1136 & 359 & 12 & 29 \\
    \multicolumn{1}{c}{} & Both Incorrect & 79 & 48 & 10 & 4 \\
    \multicolumn{1}{c}{} & Both Incorrect \& Equiv & 46 & 18 & 1 & 1 \\
    \multicolumn{1}{c}{} & Only Thinking Correct & 51 & 82 & 7 & 7 \\
    \multicolumn{1}{c}{} & Only NoThinking Correct & 53 & 11 & 1 & 0 \\
    \midrule
    \multicolumn{1}{c}{\multirow{5}{*}{R1-32B}} & Both Correct & 1192 & 369 & 13 & 28 \\
    \multicolumn{1}{c}{} & Both Incorrect & 64 & 47 & 11 & 2 \\
    \multicolumn{1}{c}{} & Both Incorrect \& Equiv & 42 & 26 & 0 & 0 \\
    \multicolumn{1}{c}{} & Only Thinking Correct & 32 & 62 & 6 & 8 \\
    \multicolumn{1}{c}{} & Only NoThinking Correct & 31 & 22 & 0 & 2 \\
    \bottomrule
    \end{tabular}}
    \caption{The analysis of consistency in the first stage of \textit{JointThinking} for main results (one group).}
    \label{table:additonal_details_of_main_results}
\end{table*}

Table \ref{table:additonal_details_of_main_results} shows the consistency analysis of answers' correctness after the first stage of \textit{JointThinking}, where reasoning large language model generates answers with thinking mode and nothinking mode in parallel. For each question, there are four possible categories: Both Correct and Both Incorrect (the answers from two different modes are both correct \& incorrect), Only Thinking Correct and Only Nothinking Correct (for two answers, only the one from thinking \& nothinking mode is correct, and the other is incorrect). These four categories are mutually exclusive and their sum is the total amount of the dataset. Both Incorrect \& Equiv is one subset of Both Incorrect which means although two answers are consistent, they are both wrong. 

In our method, when generating answers from thinking mode and nothinking mode, the answers that fall into the categories of Both Correct and Both Incorrect \& Equiv will be considered as the final answers to output, and those from Both Incorrect \& Equiv are error answers after consistency check. Others that don't pass the consistency check will be assembled into prompt with original question for the second thinking.

\begin{table*}[h]
    \centering
    \begin{tabular}{lccc}
    \toprule
    \multicolumn{1}{c}{\multirow{2}{*}{\textbf{Dataset}}} & \multicolumn{3}{c}{\textbf{Methods}} \\
    \cmidrule{2-4}
    \multicolumn{1}{c}{} & Thinking (Baseline) & JointThinking before \textless{}think\textgreater{} & JointThinking after \textless{}think\textgreater{} \\
    \midrule
    \multicolumn{4}{c}{R1-7B} \\
    \midrule
    GSM8K & 87.79 & 89.61 & 91.05 \\
    MATH500 & 87.60 & 87.20 & 87.80 \\
    AIME24 & 47.50 & 50.00 & 51.25 \\
    AMC23 & 86.88 & 87.50 & 88.44 \\
    \midrule
    \multicolumn{4}{c}{R1-14B} \\
    \midrule
    GSM8K & 89.99 & 92.34 & 93.93 \\
    MATH500 & 88.20 & 87.00 & 89.80 \\
    AIME24 & 63.75 & 63.33 & 65.42 \\
    AMC23 & 90.00 & 87.50 & 93.75 \\
    \midrule
    \multicolumn{4}{c}{Qwen3-8B} \\
    \midrule
    GSM8K & 94.69 & 95.00 & 95.00 \\
    MATH500 & 87.80 & 90.00 & 91.20 \\
    AIME24 & 59.17 & 62.08 & 71.25 \\
    AMC23 & 86.56 & 90.31 & 92.81 \\
    \midrule
    \multicolumn{4}{c}{Qwen3-14B} \\
    \midrule
    GSM8K & 94.77 & 95.38 & 95.45 \\
    MATH500 & 89.80 & 91.58 & 92.59 \\
    AIME24 & 62.92 & 61.67 & 71.67 \\
    AMC23 & 92.50 & 94.06 & 96.88 \\
    \bottomrule
    \end{tabular}
    \caption{Effect of prompt placement in the second thinking of our proposed method across different model sizes and datasets.}
    \label{table:additonal_prompt_position}
\end{table*}

\begin{table*}[h]
    \centering
    \small
    \begin{tabular}{lcccccc}
    \toprule
    \multicolumn{3}{c}{\textbf{Method}} & \textbf{GSM8K} & \textbf{MATH500} & \textbf{AIME24} & \textbf{AMC23} \\
    \midrule
    \multicolumn{1}{c}{\multirow{16}{*}{Thinking\&Thinking}} & \multicolumn{1}{c}{\multirow{2}{*}{Exp1}} & Both Correct & 1099 & 421 & 10 & 34 \\
    \cmidrule{3-7}
    \multicolumn{1}{c}{} & \multicolumn{1}{c}{} & Both Incorrect \& Equiv & 64 & 29 & 0 & 0 \\
    \cmidrule{2-7}
    \multicolumn{1}{c}{} & \multicolumn{1}{c}{\multirow{2}{*}{Exp2}} & Both Correct & 1101 & 423 & 11 & 32 \\
    \cmidrule{3-7}
    \multicolumn{1}{c}{} & \multicolumn{1}{c}{} & Both Incorrect \& Equiv & 71 & 25 & 0 & 0 \\
    \cmidrule{2-7}
    \multicolumn{1}{c}{} & \multicolumn{1}{c}{\multirow{2}{*}{Exp3}} & Both Correct & 1103 & 425 & 13 & 33 \\
    \cmidrule{3-7}
    \multicolumn{1}{c}{} & \multicolumn{1}{c}{} & Both Incorrect \& Equiv & 63 & 25 & 1 & 0 \\
    \cmidrule{2-7}
    \multicolumn{1}{c}{} & \multicolumn{1}{c}{\multirow{2}{*}{Exp4}} & Both Correct & 1107 & 425 & 12 & 33 \\
    \cmidrule{3-7}
    \multicolumn{1}{c}{} & \multicolumn{1}{c}{} & Both Incorrect \& Equiv & 60 & 24 & 0 & 0 \\
    \cmidrule{2-7}
    \multicolumn{1}{c}{} & \multicolumn{1}{c}{\multirow{2}{*}{Exp5}} & Both Correct & 1095 & 430 & 9 & 32 \\
    \cmidrule{3-7}
    \multicolumn{1}{c}{} & \multicolumn{1}{c}{} & Both Incorrect \& Equiv & 56 & 23 & 0 & 0 \\
    \cmidrule{2-7}
    \multicolumn{1}{c}{} & \multicolumn{1}{c}{\multirow{2}{*}{Exp6}} & Both Correct & 1098 & 424 & 10 & 33 \\
    \cmidrule{3-7}
    \multicolumn{1}{c}{} & \multicolumn{1}{c}{} & Both Incorrect \& Equiv & 67 & 27 & 2 & 1 \\
    \cmidrule{2-7}
    \multicolumn{1}{c}{} & \multicolumn{1}{c}{\multirow{2}{*}{Exp7}} & Both Correct & 1098 & 418 & 9 & 33 \\
    \cmidrule{3-7}
    \multicolumn{1}{c}{} & \multicolumn{1}{c}{} & Both Incorrect \& Equiv & 69 & 28 & 1 & 1 \\
    \cmidrule{2-7}
    \multicolumn{1}{c}{} & \multicolumn{1}{c}{\multirow{2}{*}{Exp8}} & Both Correct & 1106 & 420 & 10 & 34 \\
    \cmidrule{3-7}
    \multicolumn{1}{c}{} & \multicolumn{1}{c}{} & Both Incorrect \& Equiv & 67 & 28 & 0 & 0 \\
    \midrule
    \multicolumn{1}{c}{\multirow{16}{*}{Thinking\&Nothinking}} & \multicolumn{1}{c}{\multirow{2}{*}{Exp1}} & Both Correct & 1069 & 350 & 4 & 21 \\
    \cmidrule{3-7}
    \multicolumn{1}{c}{} & \multicolumn{1}{c}{} & Both Incorrect \& Equiv & 51 & 23 & 0 & 0 \\
    \cmidrule{2-7}
    \multicolumn{1}{c}{} & \multicolumn{1}{c}{\multirow{2}{*}{Exp2}} & Both Correct & 1054 & 347 & 6 & 22 \\
    \cmidrule{3-7}
    \multicolumn{1}{c}{} & \multicolumn{1}{c}{} & Both Incorrect \& Equiv & 57 & 17 & 0 & 0 \\
    \cmidrule{2-7}
    \multicolumn{1}{c}{} & \multicolumn{1}{c}{\multirow{2}{*}{Exp3}} & Both Correct & 1049 & 369 & 9 & 28 \\
    \cmidrule{3-7}
    \multicolumn{1}{c}{} & \multicolumn{1}{c}{} & Both Incorrect \& Equiv & 60 & 11 & 1 & 0 \\
    \cmidrule{2-7}
    \multicolumn{1}{c}{} & \multicolumn{1}{c}{\multirow{2}{*}{Exp4}} & Both Correct & 1074 & 361 & 4 & 23 \\
    \cmidrule{3-7}
    \multicolumn{1}{c}{} & \multicolumn{1}{c}{} & Both Incorrect \& Equiv & 50 & 18 & 0 & 0 \\
    \cmidrule{2-7}
    \multicolumn{1}{c}{} & \multicolumn{1}{c}{\multirow{2}{*}{Exp5}} & Both Correct & 1061 & 354 & 5 & 25 \\
    \cmidrule{3-7}
    \multicolumn{1}{c}{} & \multicolumn{1}{c}{} & Both Incorrect \& Equiv & 45 & 17 & 1 & 0 \\
    \cmidrule{2-7}
    \multicolumn{1}{c}{} & \multicolumn{1}{c}{\multirow{2}{*}{Exp6}} & Both Correct & 1070 & 351 & 6 & 24 \\
    \cmidrule{3-7}
    \multicolumn{1}{c}{} & \multicolumn{1}{c}{} & Both Incorrect \& Equiv & 53 & 20 & 0 & 0 \\
    \cmidrule{2-7}
    \multicolumn{1}{c}{} & \multicolumn{1}{c}{\multirow{2}{*}{Exp7}} & Both Correct & 1048 & 360 & 4 & 20 \\
    \cmidrule{3-7}
    \multicolumn{1}{c}{} & \multicolumn{1}{c}{} & Both Incorrect \& Equiv & 59 & 20 & 0 & 0 \\
    \cmidrule{2-7}
    \multicolumn{1}{c}{} & \multicolumn{1}{c}{\multirow{2}{*}{Exp8}} & Both Correct & 1065 & 355 & 8 & 22 \\
    \cmidrule{3-7}
    \multicolumn{1}{c}{} & \multicolumn{1}{c}{} & Both Incorrect \& Equiv & 53 & 21 & 1 & 0 \\
    \midrule
    \multicolumn{1}{c}{\multirow{2}{*}{Error Rate}} & \multicolumn{2}{c}{Thinking\&Thinking} & 4.90\% & 5.23\% & 1.67\% & 0.63\% \\
    \cmidrule{2-7}
    \multicolumn{1}{c}{} & \multicolumn{2}{c}{Thinking\&Nothinking} & \textbf{4.06\%} & \textbf{3.68\%} & \textbf{1.25\%} & \textbf{0.00\%} \\
    \bottomrule
    \end{tabular}
    \caption{Error rate computation details across different datasets based on the R1-7B. Thinking\&Thinking means using thinking results to calibrate thinking results, and Thinking\&Nothinking means using nothinking results to calibrate thinking results, as known as \textit{JointThinking}.}
    \label{table:additonal_error_ratio_details}
\end{table*}

\section{Additional Results of Prompt Placement}
\label{appendix:addition_results}

In order to study the effect of prompt placement in the second thinking of our proposed method, we conduct a comprehensive ablation study across different models (R1-series and Qwen3-series), sizes (7B, 8B, 14B), and different datasets (GSM8K, MATH500, AIME24, AMC23). For all running, we follow the experiment setting in Section \ref{section:experiments}, where we set the temperature to 0.6, top-p to 0.95 for the thinking mode and temperature to 0.7, top-p to 0.8 for the nothinking mode. Everything is the same, only the position of the prompt is different. \textit{JointThinking} before \textless{}think\textgreater{} means the prompt is put before the \texttt{\textless|beginning of thinking|\textgreater} special token as the normal instruction, and \textit{JointThinking} after \textless{}think\textgreater{} means the prompt is put after the \texttt{\textless|beginning of thinking|\textgreater} special token as the stater of the long cot in the reasoning processing. As shown in Table \ref{table:additonal_prompt_position}, the experiments illustrate that placing the prompt after the special token significantly outperforms the baseline and before the special token, which is one of the keys to the effectiveness of our method. We also discover that in some cases \textit{JointThinking} before \textless{}think\textgreater{} is even worse than the baseline, showing that insufficient instruction following ability limits the performance of prompt engineering.

%% file: acl_latex.bbl
\begin{thebibliography}{42}
\providecommand{\natexlab}[1]{#1}

\bibitem[{Aggarwal and Welleck(2025)}]{aggarwal2025l1}
Pranjal Aggarwal and Sean Welleck. 2025.
\newblock L1: Controlling how long a reasoning model thinks with reinforcement learning.
\newblock \emph{arXiv preprint arXiv:2503.04697}.

\bibitem[{{AI-MO}(2023)}]{amc2023}
{AI-MO}. 2023.
\newblock {Ai-mo/aimo-validation-amc}.
\newblock \url{https://huggingface.co/datasets/AI-MO/aimo-validation-amc}.
\newblock 2023.

\bibitem[{{AI-MO}(2024)}]{aime2024}
{AI-MO}. 2024.
\newblock {Ai-mo/aimo-validation-aime}.
\newblock \url{https://huggingface.co/datasets/AI-MO/aimo-validation-aime}.
\newblock 2024.

\bibitem[{Brown et~al.(2020)Brown, Mann, Ryder, Subbiah, Kaplan, Dhariwal, Neelakantan, Shyam, Sastry, Askell, Agarwal, Herbert-Voss, Krueger, Henighan, Child, Ramesh, Ziegler, Wu, Winter, Hesse, Chen, Sigler, Litwin, Gray, Chess, Clark, Berner, McCandlish, Radford, Sutskever, and Amodei}]{brown2020languagemodelsfewshotlearners}
Tom~B. Brown, Benjamin Mann, Nick Ryder, Melanie Subbiah, Jared Kaplan, Prafulla Dhariwal, Arvind Neelakantan, Pranav Shyam, Girish Sastry, Amanda Askell, Sandhini Agarwal, Ariel Herbert-Voss, Gretchen Krueger, Tom Henighan, Rewon Child, Aditya Ramesh, Daniel~M. Ziegler, Jeffrey Wu, Clemens Winter, and 12 others. 2020.
\newblock \href {https://arxiv.org/abs/2005.14165} {Language models are few-shot learners}.
\newblock \emph{Preprint}, arXiv:2005.14165.

\bibitem[{Buoso et~al.(2024)Buoso, Robinson, Averta, Torr, Franzmeyer, and Martini}]{buoso2024select2plantrainingfreeiclbasedplanning}
Davide Buoso, Luke Robinson, Giuseppe Averta, Philip Torr, Tim Franzmeyer, and Daniele~De Martini. 2024.
\newblock \href {https://arxiv.org/abs/2411.04006} {Select2plan: Training-free icl-based planning through vqa and memory retrieval}.
\newblock \emph{Preprint}, arXiv:2411.04006.

\bibitem[{Chen et~al.(2024)Chen, Xu, Liang, He, Pang, Yu, Song, Liu, Zhou, Zhang et~al.}]{chen2024not}
Xingyu Chen, Jiahao Xu, Tian Liang, Zhiwei He, Jianhui Pang, Dian Yu, Linfeng Song, Qiuzhi Liu, Mengfei Zhou, Zhuosheng Zhang, and 1 others. 2024.
\newblock Do not think that much for 2+ 3=? on the overthinking of o1-like llms.
\newblock \emph{arXiv preprint arXiv:2412.21187}.

\bibitem[{Cobbe et~al.(2021)Cobbe, Kosaraju, Bavarian, Chen, Jun, Kaiser, Plappert, Tworek, Hilton, Nakano et~al.}]{cobbe2021training}
Karl Cobbe, Vineet Kosaraju, Mohammad Bavarian, Mark Chen, Heewoo Jun, Lukasz Kaiser, Matthias Plappert, Jerry Tworek, Jacob Hilton, Reiichiro Nakano, and 1 others. 2021.
\newblock Training verifiers to solve math word problems.
\newblock \emph{arXiv preprint arXiv:2110.14168}.

\bibitem[{Cuadron et~al.(2025)Cuadron, Li, Ma, Wang, Wang, Zhuang, Liu, Schroeder, Xia, Mao et~al.}]{cuadron2025danger}
Alejandro Cuadron, Dacheng Li, Wenjie Ma, Xingyao Wang, Yichuan Wang, Siyuan Zhuang, Shu Liu, Luis~Gaspar Schroeder, Tian Xia, Huanzhi Mao, and 1 others. 2025.
\newblock The danger of overthinking: Examining the reasoning-action dilemma in agentic tasks.
\newblock \emph{arXiv preprint arXiv:2502.08235}.

\bibitem[{Ge et~al.(2025)Ge, Liu, Wang, Mei, Chen, Bi, and Cheng}]{ge2025innate}
Yuyao Ge, Shenghua Liu, Yiwei Wang, Lingrui Mei, Lizhe Chen, Baolong Bi, and Xueqi Cheng. 2025.
\newblock Innate reasoning is not enough: In-context learning enhances reasoning large language models with less overthinking.
\newblock \emph{arXiv preprint arXiv:2503.19602}.

\bibitem[{Guo et~al.(2025)Guo, Yang, Zhang, Song, Zhang, Xu, Zhu, Ma, Wang, Bi et~al.}]{guo2025deepseek}
Daya Guo, Dejian Yang, Haowei Zhang, Junxiao Song, Ruoyu Zhang, Runxin Xu, Qihao Zhu, Shirong Ma, Peiyi Wang, Xiao Bi, and 1 others. 2025.
\newblock Deepseek-r1: Incentivizing reasoning capability in llms via reinforcement learning.
\newblock \emph{arXiv preprint arXiv:2501.12948}.

\bibitem[{Hosseini et~al.(2022)Hosseini, Vani, Bahdanau, Sordoni, and Courville}]{hosseini-etal-2022-compositional}
Arian Hosseini, Ankit Vani, Dzmitry Bahdanau, Alessandro Sordoni, and Aaron Courville. 2022.
\newblock \href {https://doi.org/10.18653/v1/2022.blackboxnlp-1.22} {On the compositional generalization gap of in-context learning}.
\newblock In \emph{Proceedings of the Fifth BlackboxNLP Workshop on Analyzing and Interpreting Neural Networks for NLP}, pages 272--280, Abu Dhabi, United Arab Emirates (Hybrid). Association for Computational Linguistics.

\bibitem[{Lampinen et~al.(2025)Lampinen, Chaudhry, Chan, Wild, Wan, Ku, Bornschein, Pascanu, Shanahan, and McClelland}]{lampinen2025generalization}
Andrew~K Lampinen, Arslan Chaudhry, Stephanie~CY Chan, Cody Wild, Diane Wan, Alex Ku, J{\"o}rg Bornschein, Razvan Pascanu, Murray Shanahan, and James~L McClelland. 2025.
\newblock On the generalization of language models from in-context learning and finetuning: a controlled study.
\newblock \emph{arXiv preprint arXiv:2505.00661}.

\bibitem[{Li et~al.(2024)Li, Jing, Li, Zhai, Wu, and Jia}]{li-etal-2024-context}
Chuanhao Li, Chenchen Jing, Zhen Li, Mingliang Zhai, Yuwei Wu, and Yunde Jia. 2024.
\newblock \href {https://doi.org/10.18653/v1/2024.emnlp-main.996} {In-context compositional generalization for large vision-language models}.
\newblock In \emph{Proceedings of the 2024 Conference on Empirical Methods in Natural Language Processing}, pages 17954--17966, Miami, Florida, USA. Association for Computational Linguistics.

\bibitem[{Li et~al.(2025{\natexlab{a}})Li, Lv, Shao, Ma, Li, Zheng, Qiu, and Guo}]{li2025fastmcts}
Peiji Li, Kai Lv, Yunfan Shao, Yichuan Ma, Linyang Li, Xiaoqing Zheng, Xipeng Qiu, and Qipeng Guo. 2025{\natexlab{a}}.
\newblock Fastmcts: A simple sampling strategy for data synthesis.
\newblock \emph{arXiv preprint arXiv:2502.11476}.

\bibitem[{Li et~al.(2025{\natexlab{b}})Li, Yu, Zhang, Chen, Zhang, Zhuang, Sadagopan, and Beniwal}]{li2025thinking}
Xiaomin Li, Zhou Yu, Zhiwei Zhang, Xupeng Chen, Ziji Zhang, Yingying Zhuang, Narayanan Sadagopan, and Anurag Beniwal. 2025{\natexlab{b}}.
\newblock When thinking fails: The pitfalls of reasoning for instruction-following in llms.
\newblock \emph{arXiv preprint arXiv:2505.11423}.

\bibitem[{Liao et~al.(2025)Liao, Xu, Dong, Li, Monz, Savarese, Sahoo, and Xiong}]{liao2025reward}
Baohao Liao, Yuhui Xu, Hanze Dong, Junnan Li, Christof Monz, Silvio Savarese, Doyen Sahoo, and Caiming Xiong. 2025.
\newblock Reward-guided speculative decoding for efficient llm reasoning.
\newblock \emph{arXiv preprint arXiv:2501.19324}.

\bibitem[{Lightman et~al.(2023)Lightman, Kosaraju, Burda, Edwards, Baker, Lee, Leike, Schulman, Sutskever, and Cobbe}]{lightman2023let}
Hunter Lightman, Vineet Kosaraju, Yuri Burda, Harrison Edwards, Bowen Baker, Teddy Lee, Jan Leike, John Schulman, Ilya Sutskever, and Karl Cobbe. 2023.
\newblock Let's verify step by step.
\newblock In \emph{The Twelfth International Conference on Learning Representations}.

\bibitem[{Lou et~al.(2025)Lou, Sun, Liang, Qu, Shen, Wang, Li, Yang, and Wu}]{lou2025adacot}
Chenwei Lou, Zewei Sun, Xinnian Liang, Meng Qu, Wei Shen, Wenqi Wang, Yuntao Li, Qingping Yang, and Shuangzhi Wu. 2025.
\newblock Adacot: Pareto-optimal adaptive chain-of-thought triggering via reinforcement learning.
\newblock \emph{arXiv preprint arXiv:2505.11896}.

\bibitem[{Ma et~al.(2025{\natexlab{a}})Ma, He, Snell, Griggs, Min, and Zaharia}]{ma2025reasoning}
Wenjie Ma, Jingxuan He, Charlie Snell, Tyler Griggs, Sewon Min, and Matei Zaharia. 2025{\natexlab{a}}.
\newblock Reasoning models can be effective without thinking.
\newblock \emph{arXiv preprint arXiv:2504.09858}.

\bibitem[{Ma et~al.(2025{\natexlab{b}})Ma, Wan, Yu, Fang, and Wang}]{ma2025cot}
Xinyin Ma, Guangnian Wan, Runpeng Yu, Gongfan Fang, and Xinchao Wang. 2025{\natexlab{b}}.
\newblock Cot-valve: Length-compressible chain-of-thought tuning.
\newblock \emph{arXiv preprint arXiv:2502.09601}.

\bibitem[{Meade et~al.(2023)Meade, Gella, Hazarika, Gupta, Jin, Reddy, Liu, and Hakkani-T{\"u}r}]{meade2023using}
Nicholas Meade, Spandana Gella, Devamanyu Hazarika, Prakhar Gupta, Di~Jin, Siva Reddy, Yang Liu, and Dilek Hakkani-T{\"u}r. 2023.
\newblock Using in-context learning to improve dialogue safety.
\newblock \emph{arXiv preprint arXiv:2302.00871}.

\bibitem[{OpenAI(2024)}]{OpenAI2024IntroducingO3}
OpenAI. 2024.
\newblock Introducing o3 and o4 mini.
\newblock \url{https://openai.com/index/introducing-o3-and-o4-mini/}.
\newblock Accessed: 2025-06-30.

\bibitem[{Peng et~al.(2024)Peng, Ding, Yuan, Liu, Zhang, Ouyang, and Tao}]{peng-etal-2024-revisiting}
Keqin Peng, Liang Ding, Yancheng Yuan, Xuebo Liu, Min Zhang, Yuanxin Ouyang, and Dacheng Tao. 2024.
\newblock \href {https://doi.org/10.18653/v1/2024.acl-long.492} {Revisiting demonstration selection strategies in in-context learning}.
\newblock In \emph{Proceedings of the 62nd Annual Meeting of the Association for Computational Linguistics (Volume 1: Long Papers)}, pages 9090--9101, Bangkok, Thailand. Association for Computational Linguistics.

\bibitem[{Rein et~al.(2024)Rein, Hou, Stickland, Petty, Pang, Dirani, Michael, and Bowman}]{rein2024gpqa}
David Rein, Betty~Li Hou, Asa~Cooper Stickland, Jackson Petty, Richard~Yuanzhe Pang, Julien Dirani, Julian Michael, and Samuel~R Bowman. 2024.
\newblock Gpqa: A graduate-level google-proof q\&a benchmark.
\newblock In \emph{First Conference on Language Modeling}.

\bibitem[{Shao et~al.(2024)Shao, Wang, Zhu, Xu, Song, Bi, Zhang, Zhang, Li, Wu et~al.}]{shao2024deepseekmath}
Zhihong Shao, Peiyi Wang, Qihao Zhu, Runxin Xu, Junxiao Song, Xiao Bi, Haowei Zhang, Mingchuan Zhang, YK~Li, Y~Wu, and 1 others. 2024.
\newblock Deepseekmath: Pushing the limits of mathematical reasoning in open language models.
\newblock \emph{arXiv preprint arXiv:2402.03300}.

\bibitem[{Shi et~al.(2024)Shi, Wei, Xu, and Liang}]{shi2024largerlanguagemodelsincontext}
Zhenmei Shi, Junyi Wei, Zhuoyan Xu, and Yingyu Liang. 2024.
\newblock \href {https://arxiv.org/abs/2405.19592} {Why larger language models do in-context learning differently?}
\newblock \emph{Preprint}, arXiv:2405.19592.

\bibitem[{Snell et~al.(2024)Snell, Lee, Xu, and Kumar}]{snell2024scaling}
Charlie Snell, Jaehoon Lee, Kelvin Xu, and Aviral Kumar. 2024.
\newblock Scaling llm test-time compute optimally can be more effective than scaling model parameters.
\newblock \emph{arXiv preprint arXiv:2408.03314}.

\bibitem[{Sun et~al.(2024)Sun, Haider, Zhang, Yang, Qiu, Yin, Wang, Bartlett, and Zanette}]{sun2024fast}
Hanshi Sun, Momin Haider, Ruiqi Zhang, Huitao Yang, Jiahao Qiu, Ming Yin, Mengdi Wang, Peter Bartlett, and Andrea Zanette. 2024.
\newblock Fast best-of-n decoding via speculative rejection.
\newblock \emph{arXiv preprint arXiv:2410.20290}.

\bibitem[{Tang and Dong(2024)}]{tang2024demonstrationnotebookfindingsuited}
Yiming Tang and Bin Dong. 2024.
\newblock \href {https://arxiv.org/abs/2406.10878} {Demonstration notebook: Finding the most suited in-context learning example from interactions}.
\newblock \emph{Preprint}, arXiv:2406.10878.

\bibitem[{Wan et~al.(2025)Wan, Wang, Xu, Yin, Song, and Sun}]{wan2025adapthink}
Xu~Wan, Wei Wang, Wenyue Xu, Wotao Yin, Jie Song, and Mingyang Sun. 2025.
\newblock Adapthink: Adaptive thinking preferences for reasoning language model.
\newblock \emph{arXiv preprint arXiv:2506.18237}.

\bibitem[{Wang et~al.(2025{\natexlab{a}})Wang, Wu, Yuan, Cai, Li, and Jia}]{wang2025demonstrationselectionincontextlearning}
Xubin Wang, Jianfei Wu, Yichen Yuan, Deyu Cai, Mingzhe Li, and Weijia Jia. 2025{\natexlab{a}}.
\newblock \href {https://arxiv.org/abs/2412.03966} {Demonstration selection for in-context learning via reinforcement learning}.
\newblock \emph{Preprint}, arXiv:2412.03966.

\bibitem[{Wang et~al.(2025{\natexlab{b}})Wang, Zhang, Huang, Yang, Zhang, Huang, and Wang}]{wang2025sampling}
Yiming Wang, Pei Zhang, Siyuan Huang, Baosong Yang, Zhuosheng Zhang, Fei Huang, and Rui Wang. 2025{\natexlab{b}}.
\newblock Sampling-efficient test-time scaling: Self-estimating the best-of-n sampling in early decoding.
\newblock \emph{arXiv preprint arXiv:2503.01422}.

\bibitem[{Wang et~al.(2024)Wang, Ma, Zhang, Ni, Chandra, Guo, Ren, Arulraj, He, Jiang et~al.}]{wang2024mmlu}
Yubo Wang, Xueguang Ma, Ge~Zhang, Yuansheng Ni, Abhranil Chandra, Shiguang Guo, Weiming Ren, Aaran Arulraj, Xuan He, Ziyan Jiang, and 1 others. 2024.
\newblock Mmlu-pro: A more robust and challenging multi-task language understanding benchmark.
\newblock In \emph{The Thirty-eight Conference on Neural Information Processing Systems Datasets and Benchmarks Track}.

\bibitem[{Wei et~al.(2022)Wei, Wang, Schuurmans, Bosma, Xia, Chi, Le, Zhou et~al.}]{wei2022chain}
Jason Wei, Xuezhi Wang, Dale Schuurmans, Maarten Bosma, Fei Xia, Ed~Chi, Quoc~V Le, Denny Zhou, and 1 others. 2022.
\newblock Chain-of-thought prompting elicits reasoning in large language models.
\newblock \emph{Advances in neural information processing systems}, 35:24824--24837.

\bibitem[{Wei et~al.(2023)Wei, Hou, Lampinen, Chen, Huang, Tay, Chen, Lu, Zhou, Ma et~al.}]{wei2023symbol}
Jerry Wei, Le~Hou, Andrew Lampinen, Xiangning Chen, Da~Huang, Yi~Tay, Xinyun Chen, Yifeng Lu, Denny Zhou, Tengyu Ma, and 1 others. 2023.
\newblock Symbol tuning improves in-context learning in language models.
\newblock \emph{arXiv preprint arXiv:2305.08298}.

\bibitem[{Wies et~al.(2023)Wies, Levine, and Shashua}]{wies2023learnability}
Noam Wies, Yoav Levine, and Amnon Shashua. 2023.
\newblock The learnability of in-context learning.
\newblock \emph{Advances in Neural Information Processing Systems}, 36:36637--36651.

\bibitem[{Yang et~al.(2025)Yang, Li, Yang, Zhang, Hui, Zheng, Yu, Gao, Huang, Lv et~al.}]{yang2025qwen3}
An~Yang, Anfeng Li, Baosong Yang, Beichen Zhang, Binyuan Hui, Bo~Zheng, Bowen Yu, Chang Gao, Chengen Huang, Chenxu Lv, and 1 others. 2025.
\newblock Qwen3 technical report.
\newblock \emph{arXiv preprint arXiv:2505.09388}.

\bibitem[{Yu et~al.(2024)Yu, Xu, Weston, and Kulikov}]{yu2024distilling}
Ping Yu, Jing Xu, Jason Weston, and Ilia Kulikov. 2024.
\newblock Distilling system 2 into system 1.
\newblock \emph{arXiv preprint arXiv:2407.06023}.

\bibitem[{Yu et~al.(2025)Yu, Wu, Zhao, Cohan, and Zhang}]{yu2025z1}
Zhaojian Yu, Yinghao Wu, Yilun Zhao, Arman Cohan, and Xiao-Ping Zhang. 2025.
\newblock Z1: Efficient test-time scaling with code.
\newblock \emph{arXiv preprint arXiv:2504.00810}.

\bibitem[{Zhang et~al.(2025{\natexlab{a}})Zhang, Lin, Hou, Feng, and Li}]{zhang2025adaptthink}
Jiajie Zhang, Nianyi Lin, Lei Hou, Ling Feng, and Juanzi Li. 2025{\natexlab{a}}.
\newblock Adaptthink: Reasoning models can learn when to think.
\newblock \emph{arXiv preprint arXiv:2505.13417}.

\bibitem[{Zhang et~al.(2025{\natexlab{b}})Zhang, Cao, Wei, You, and Ding}]{zhang2025promptdesignmattersworks}
Xiang Zhang, Juntai Cao, Jiaqi Wei, Chenyu You, and Dujian Ding. 2025{\natexlab{b}}.
\newblock \href {https://arxiv.org/abs/2503.10084} {Why prompt design matters and works: A complexity analysis of prompt search space in llms}.
\newblock \emph{Preprint}, arXiv:2503.10084.

\bibitem[{Zhang et~al.(2025{\natexlab{c}})Zhang, Wang, Li, Xue, Guan, Xu, Zou, Yu, and Cui}]{zhang2025understanding}
Xingxuan Zhang, Haoran Wang, Jiansheng Li, Yuan Xue, Shikai Guan, Renzhe Xu, Hao Zou, Han Yu, and Peng Cui. 2025{\natexlab{c}}.
\newblock \href {https://openreview.net/forum?id=yOhNLIqTEF} {Understanding the generalization of in-context learning in transformers: An empirical study}.
\newblock In \emph{The Thirteenth International Conference on Learning Representations}.

\end{thebibliography}
